\documentclass[pdflatex,sn-mathphys-num]{sn-jnl}% Math and Physical Sciences Numbered Reference Style
\usepackage{graphicx}%
\usepackage{multirow}%
\usepackage{amsmath,amssymb,amsfonts}%
\usepackage{amsthm}%
\usepackage{mathrsfs}%
\usepackage[title]{appendix}%
\usepackage{xcolor}%
\usepackage{textcomp}%
\usepackage{manyfoot}%
\usepackage{booktabs}%
\usepackage{algorithm}%
\usepackage{algorithmicx}%
\usepackage{algpseudocode}%
\usepackage{listings}%
\usepackage{enumitem}
\usepackage{float}

\usepackage[version=3]{mhchem} % Formula subscripts using \ce{}
\usepackage{xcolor}
\usepackage{microtype}
\usepackage{amsfonts}
\usepackage[T1]{fontenc}
\usepackage{lmodern}
\usepackage{hyperref}

\usepackage{cleveref}
\usepackage{comment}
\usepackage{caption}

\usepackage{booktabs}
\usepackage{siunitx}
\usepackage{tikz}
\usetikzlibrary{positioning}
\usetikzlibrary{arrows.meta}

\newcommand{\red}[1]{\textcolor{black}{#1}}

\newcommand{\datasetstats}[4]{%
  \textit{Dataset size:} #1; %
  \textit{dimensionality:} #2; %
  \textit{target:} #3; %
  \textit{objective:} #4.%
}

\theoremstyle{thmstyleone}%
\theoremstyle{thmstyletwo}%

\theoremstyle{thmstylethree}%

\AtBeginDocument{\addtolength{\textheight}{36pt}} % try 24–48pt

\begin{document}

\title[Article Title]{Frugal Bayesian Optimization: Scalable Surrogates for Data- and Resource-Limited Discovery}


\author*[1]{\fnm{Panagiotis} \sur{Krokidas}}\email{p.krokidas@iit.demokritos.gr}

\author[1,2]{\fnm{Christoforos} \sur{Rekatsinas}}\email{crek@iit.demokritos.gr}
\author[1,3]{\fnm{Vassilis} \sur{Sioros}}\email{vsioros@iit.demokritos.gr}

\author[4]{\fnm{Grigorios M.} \sur{Chatziathanasiou}}\email{grigorischatz@hmu.gr}

\author[5,6]{\fnm{Efi-Maria} \sur{Papia}}\email{e.papia@inn.demokritos.gr}

\author[1,7]{\fnm{George} \sur{Giannakopoulos}}\email{ggianna@iit.demokritos.gr}

\affil*[1]{\orgdiv{Institute of Informatics and Telecommunications}, \orgname{National Centre for Scientific Research "Demokritos"}, \orgaddress{
% \street{},
\city{Agia Paraskevi},
% \postcode{}, \state{},
\country{Greece}}}

\affil[2]{\orgdiv{Department of Mechanical Engineering and Aeronautics}, \orgname{University of Patras}, \orgaddress{\city{Patras} \country{Greece}}}

\affil[3]{\orgdiv{Department of Informatics and Telecommunications}, \orgname{National and Kapodistrian University of Athens}, \orgaddress{ \city{Athens}, \country{Greece}}}

\affil[4]{\orgdiv{School of Mechanical Engineering}, \orgname{Hellenic Mediterranean
University}, \orgaddress{\city{Heraklion}, \state{Crete}, \country{Greece}}}

\affil*[5]{\orgdiv{Institute of Nanoscience and Nanotechnology}, \orgname{National Centre for Scientific Research "Demokritos"}, \orgaddress{\city{Agia Paraskevi}, \country{Greece}}}

\affil[6]{\orgdiv{Department of Physics}, \orgname{National and Kapodistrian University of Athens}, \orgaddress{\city{Athens}, \country{Greece}}}

\affil[7]{\orgname{SciFY PNPC}, \orgaddress{\city{Agia Paraskevi}, \country{Greece}}}


\abstract{
Bayesian Optimization (BO) is widely adopted for data-efficient optimization in scientific and engineering applications, yet its computational cost is rarely evaluated alongside optimization performance. Here we present a systematic, compute-aware study of BO that redefines surrogate evaluation along two axes: optimization quality and computational frugality. Across eight benchmark functions and nine real-world datasets spanning materials science, mechanics, robotics, chemistry, and machine learning, we benchmark four surrogate models—Gaussian Processes, Random Forests, NGBoost, and Bayesian Adaptive Spline Surfaces. We show that Gaussian Process–based BO consistently incurs the highest time and memory overhead without delivering superior optimization or sample efficiency. In contrast, scalable alternatives achieve equal or better performance at a fraction of the computational cost. Motivated by these findings, we introduce a surrogate-recommendation framework that predicts the most suitable BO surrogate from inexpensive dataset characteristics. Together, these results establish FruBO as a reproducible, compute-aware baseline for Bayesian Optimization and provide practical guidance for surrogate selection under limited computational and experimental budgets.
}




\maketitle

With the advent of machine learning (ML), data-driven methods have become integral to daily life. Production systems (recommendation engines, spam filters, driver-assistance) thrive on abundant data, enabling models to interpolate within well-covered distributions. Scientific discovery faces a different regime: data are scarce, and the goal is not to model the average case but to extrapolate toward rare, exceptional solutions. Yet extrapolation is notoriously brittle for models trained on random samples that under-cover the regions of highest interest. 

To move beyond passive sampling, the community has revisited active learning (AL)~\cite{Settles2009active}: rather than consuming data indiscriminately, the model proposes which sample to acquire next (e.g., a new experiment), guided by informativeness criteria. While AL can yield good predictors with fewer measurements, its iterative loop is typically geared toward representing the space broadly. This does not guarantee accuracy precisely where scientists care most: near the global optimum or within a top-$N$ subset of high-value candidates. In practice, the scientist’s question is rarely ``what happens across the entire design space?'' but rather ``where is the best solution in the unexplored design space?'' This shifts the problem from broad representation to targeted optimization.

Bayesian Optimization (BO)~\cite{Shahriari2016bo} addresses this optimization-centric goal by coupling a surrogate model with an acquisition rule to navigate design spaces efficiently. In materials and chemistry, BO has steered searches toward high-performing regions, e.g., for material and drug discovery~\cite{Chitturi2024,Wu2024,Loutas2025,Krokidas2025}. However, standard BO pipelines rely on Gaussian processes (GPs), whose per-iteration updates involve covariance factorizations with $\mathcal{O}(N^3)$ time and memory that grows with $N$ (often effectively $\mathcal{O}(N^2)$ due to kernel matrices)~\cite{Rasmussen2006, Siemenn2023_}. As acquisitions accumulate, wall-clock time and memory can become prohibitive on modest hardware, pushing users toward specialized GPUs or HPC clusters. Although BO often outperforms state-of-the-art non-BO optimizers, such as evolutionary strategies (e.g., genetic algorithm, CMA-ES) and particle swarm optimization, in terms of sample efficiency at small evaluation budgets~\cite{Tani2021,Tani2024,Santoni2024}, the GP overhead can dominate as $N$ grows. This is especially true when objective evaluations are inexpensive or highly parallelizable, turning surrogate updates into the bottleneck and limiting BO’s practical use over other optimizers~\cite{Tani2024,Santoni2024}.

A large body of work has sought to mitigate the scaling bottlenecks of GP-based BO—via bounded training sets and zooming strategies that progressively narrow the search space~\cite{Siemenn2023}, focalized sparse-GP objectives that concentrate learning on high-acquisition regions~\cite{Wei2024}, goal-aware acquisition functions that optimize for user-defined subsets rather than the global optimum~\cite{Chitturi2024}, fixed-memory buffer architectures that cap kernel growth~\cite{Chang2023}, and even hardware acceleration through in-memory or neuromorphic computing~\cite{Lin2024}. These approaches can substantially delay the onset of cubic cost, but once we commit to GP posteriors we inevitably inherit additional machinery—sparse approximations, variational bounds, or posterior sampling—that increases both methodological and engineering complexity. Such layers of approximation often require expert tuning and careful calibration of hyperparameters, while still failing to eliminate the fundamental memory growth of GP kernels. This motivates exploring surrogates that render BO’s cost nearly $\mathcal{O}(N)$ in time with respect to the number of acquisitions $N$, while keeping memory usage effectively constant (bounded) in $N$.

Recent studies have indeed begun to explore non-GP surrogates. Gradient-informed Bayesian neural networks trained with stochastic-gradient MCMC can incorporate derivative information and improve sample efficiency, yet explicit wall-clock and memory comparisons against GPs are rarely quantified~\cite{Makrygiorgos2025}. The only compute evidence provided is a report of essentially identical training times across three functions—McCormick (2D), Rosenbrock (4D), and Hartmann–6—of about $23$\,s per training run, indicating stability with respect to dimensionality but no scaling study versus dataset size, nor memory measurements or head-to-head comparisons with GPs. Broader benchmarks examining finite- and infinite-width BNNs, Laplace approximations, deep kernel learning (DKL), and ensemble models show that GPs are not uniquely superior for identifying optima; performance is problem-dependent, with HMC-based BNNs excelling in some regimes and DKL remaining consistently competitive~\cite{Li2024}. However, these studies prioritize optimization outcomes and do not profile wall-clock time or memory usage. \citet{Lakshminarayanan2016} study Mondrian forests (MFs) as BO surrogates: MFs extend decision forests with a hierarchical Gaussian prior to yield calibrated uncertainty and efficient online updates; on four BO benchmarks (Branin, Hartmann–6, SVM-grid, LDA-grid) MF–UCB matches or exceeds SMAC (RF), and on a large flight-delay regression task MF attains better NLPD than a strong GP approximation (rBCM). The study, however, reports no wall-clock or memory profiles (compute is discussed via $\mathcal{O}(N\log N)$ vs.\ $\mathcal{O}(N^3)$ arguments rather than measurements), and the BO comparison is against Random Forest (RF) rather than GP. In a representative low-data setting, \citet{Tom2023} runs discrete-library BO across six small molecular datasets (150–1{,}870 candidates) using GP surrogates (including a Tanimoto-kernel variant), NGBoost, SNGP, BNN, and GNNGP, but reports only optimization/calibration results without wall-clock or memory measurements. Likewise, Bayesian model averaging frameworks that integrate non-GP surrogates such as BART and BMARS demonstrate strong optimization outcomes on practical, discrete candidate sets~\cite{Lei2021}, again without compute or memory profiling. An exception on the compute side is the DNGO study of \citet{Snoek2015}, which includes a direct runtime-scaling plot against a GP baseline (Spearmint) on the Hartmann–6 benchmark—showing roughly linear growth for DNGO versus much steeper increases for the GP—yet this is limited to a single benchmark and does not report memory.

Altogether, the field still lacks a systematic, quantitative comparison that elevates both optimization quality and resource cost to first-class metrics, leaving open the central question of whether alternative surrogates can deliver not only accuracy but also genuine efficiency gains over GP-based BO. At the same time, frugality matters: data centers and AI form an energy-hungry duo---DC usage is large and rising~\cite{Andrae2015,Yang2024} and AI's footprint keeps growing despite efficiency gains, with inference up to $\sim$60\%~\cite{Patterson2022,Tripp2024,DeVries2023}. Hardware demand and costs are climbing in parallel---driven by AI workloads and acute GPU scarcity~\cite{Mehta2023}---tightening compute budgets in research settings. Even if scientific ML workloads are modest compared to frontier LLMs, adopting a resource-aware mindset (minimizing wall-clock time and memory, and avoiding unnecessary GPUs/HPC) should be standard practice.
ML/AI is rapidly being deployed in domains central to net-zero goals (renewable-ready power grids \cite{Steven2026}, transport and logistics \cite{APMS2025_PartIV}, buildings and industrial operations \cite{Rahman2025}, land-use and agriculture~\cite{Naturinda2025}, and accelerated discovery of materials for batteries \cite{Xin2026}, renewables \cite{Bai2025}, Cooling/Dehumidification \cite{Liu2025} and $\mathrm{CO}_2$ emissions reduction \cite{Bose2023}) where it can cut emissions; yet a growing discussion warns that escalating AI compute in net-zero research can offset these gains and even delay net-zero timelines~\cite{Luers2024, Luers2025}. This tension further motivates resource-aware methods in scientific ML.
Historically, GP-based BO has been the default; here we demonstrate that alternative surrogates can replace GPs to deliver comparable or better optimization while reducing compute and memory, enabling more efficient sampling.

FruBO framework is a redefinition of how BO should be evaluated. We reframe BO as a two-axis problem (balancing sampling efficiency with computational cost) and provide the first systematic, quantitative study across surrogate models that exposes this trade-off. Within this perspective, efficiency is not measured solely by sample or iteration count, but by the joint behavior of optimization quality and computational scaling (time, memory) as acquisitions grow. We benchmark four surrogates: a GP baseline and three scalable alternatives---Random Forests (RF), Bayesian Adaptive Spline Surfaces (BASS)~\cite{Francom2020}, and NGBoost (NGB)~\cite{Duan2020}. FruBO offers a practical and transparent way to measure and visualize this balance, establishing reference behavior that future BO research can build on. By standardizing this paired evaluation protocol and releasing it as open-source software, we aim to encourage a culture of compute-aware benchmarking in scientific machine learning and to guide the selection and development of surrogate models under realistic resource constraints.

We evaluate across two classes of datasets. First, eight well-known benchmark functions which offer controlled landscapes with known optima: Rastrigin, Ackley, Schwefel, Michalewicz, Schaffer~7, Styblinski--Tang, Weierstrass, and Expanded Schaffer~F6. Second, 9 domain-relevant problems that differ in representation, smoothness, dimensionality, and noise: (i) a approx. 70{,}000 COF candidates dataset for maximizing the methane deliverable capacity; (ii) a 400{,}000-candidate moiré-pattern system of six-layer stacks targeting near-uniform pore-size distributions; (iii) an optimal-control task based on LunarLanderContinuous-v3 reformulated for BO by parameterizing action schedules and using episode return as the black-box objective; (iv) a large-scale molecular screening for therapeutic relevance, looking for the molecule with the lowest docking energy among 400,000 candidates from a large-scale Docking Dataset; (v) $\sim$50{,}000 modeled high-pressure tank designs to maximize pressure thresholds; (vi) the 130,000 small molecules QM9 dataset to maximize the HUMO-LUOMO gap value; (vii) an approx. 580,000 candidates dataset of a multi-layer approaximation of a c. elegance model towards minimizing its stiffness; (viii) a neural-network hyperparameter optimization on a scientific task; (ix) and an electromechanical damping towards minimizing structure vibrations across multiple frequencies.

Within this framework, we find that non-GP surrogates consistently match or surpass GP-based BO in terms of optimization performance (best-found value and recall@100 as a function of samples), while offering markedly superior computational efficiency (best-found value and recall@100 as a function of compute time). The compute profile differs sharply across models: GP training time grows superlinearly with the number of acquisitions and memory accumulates with each update, whereas RF, BASS, and NGB exhibit nearly linear time scaling with effectively bounded memory usage. In practical terms, BO runs with GP surrogates can terminate prematurely on standard workstations due to memory pressure, even in settings where the objective evaluations themselves are inexpensive. By contrast, the non-GP surrogates remain stable and efficient throughout. These findings demonstrate that the surrogate choice is not a secondary modeling detail but a primary determinant of computational scalability. As acquisition counts increase, the frugality-oriented surrogates deliver equal or better optimization quality at a fraction of the computational cost.

Motivated by these results, we take a further step and formalize the selection of surrogates through classifier. Using a small set of easily obtainable dataset characteristics, such as dataset size, dimensionality, fractal dimension, and a lightweight estimate of the target variance, we train a multi-output model that predicts the expected ranking of the four surrogates under either compute-time–limited or sample-acquisition–limited regimes. This approach transforms empirical observations into a practical tool: given a new dataset, users can obtain an informed recommendation for the surrogate model most likely to balance optimization performance and computational frugality. In this way, FruBO not only quantifies the trade-offs inherent in surrogate choice but also operationalizes them, providing an accessible and compute-aware BO strategy for researchers working under constrained data or hardware budgets.

In summary, this study combines three complementary elements: a systematic comparison of BO surrogate models across diverse problem classes, a compute-aware evaluation protocol that jointly assesses optimization quality and computational scaling, and a lightweight surrogate-recommendation model that capitalizes on the resulting empirical structure. By linking surrogate benchmarking, evaluation, and selection within a single framework, FruBO provides a practical methodology for deploying Bayesian Optimization in settings where both data and computational resources are constrained.

\section{Results}\label{sec2}
\subsection{The premise}\label{subsec2}
The premise of this work is that researchers typically operate under fixed experimental or computational budgets while seeking high-performing solutions within large, complex design spaces (for example, materials screening under limited simulation time or laboratory experiments constrained by cost). In our setting, this budget is fixed at $N=1000$ new acquisitions (e.g., synthesizing a new material and measuring its performance), which is typically much smaller than the total number of candidates in the design space ($N \ll N'$, where $N'$ denotes the full set of available materials or design options). In addition to optimization outcomes, we explicitly track the computational costs incurred during the BO process. For each surrogate, we report: (i) training time as a function of the number of acquisitions, reflecting how wall-clock cost scales over the course of the loop, and (ii) GPU memory consumption as a function of the number of acquisitions, reflecting whether memory usage remains bounded or grows with dataset size. These two dimensions provide a direct measure of scalability and practical feasibility on standard GPU-equipped workstations, complementing the evaluation of optimization performance. The surrogate models are the standard GP, RF, NGBoost and BASS, and more information for their implementation in our work can be found in the methods section.

\subsection{Benchmark functions}
 
\begin{figure}[H]
    \centering
    \includegraphics[width=0.99\linewidth]{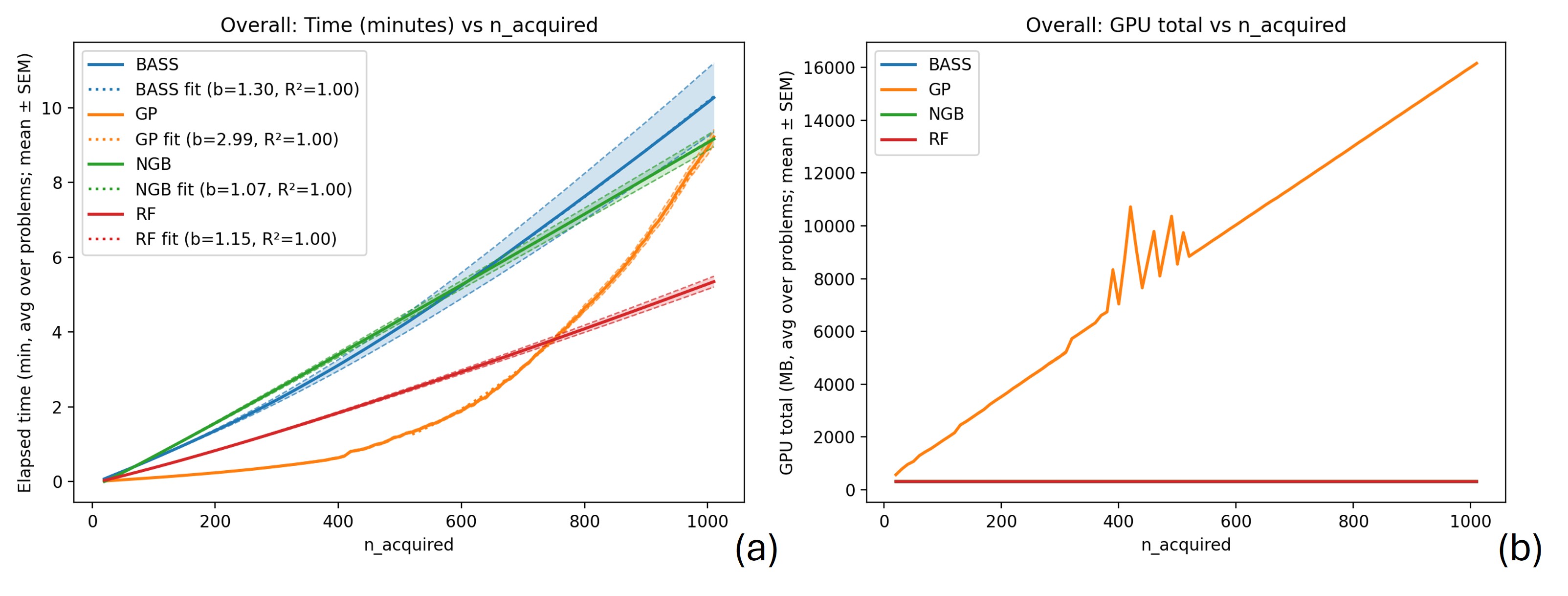}
    \caption{Aggregated BO performance across nine benchmark functions. (a) wall-clock training time vs.\ acquisitions, and (b) GPU memory vs.\ acquisitions. Curves are averaged over functions and BO seeds; shaded regions indicate variability.}
    \label{fig:functions_overallComputeGPU}
\end{figure}

Figure~\ref{fig:functions_overallComputeGPU} reports the aggregated compute-performance results (wall-clock time and GPU memory) across all eight benchmark functions, each averaged over 20 BO runs. Shaded regions indicate variability across BO seeds. Because all benchmark functions share identical computational characteristics—namely the same dataset size (up to 1{,}000 acquired samples) and the same input dimensionality (4D)—this aggregation is meaningful and provides a reliable overview of surrogate-model scaling behavior as the number of BO acquisitions increases. As expected, GP-based BO exhibits the steepest growth in compute cost. The wall-clock training time for GP increases superlinearly with the number of acquired samples, consistent with the well-known \(\mathcal{O}(n^3)\) complexity of Gaussian Process regression. The GPU footprint of GP also grows approximately linearly, reflecting the increasing cost of kernel-matrix operations. In contrast, the alternative surrogates, RF, NGB, and BASS, display markedly different scaling. Their wall-clock training time increases only linearly with the number of acquisitions, and their GPU memory usage remains minimal throughout the entire BO trajectory. Importantly, the GPU cost of these models does not increase as more samples are acquired, a consistent and striking result that also appears in the real-world datasets analyzed later. These findings highlight a key practical insight: in settings with larger datasets or higher-dimensional feature spaces (beyond the 4D benchmark functions used here), replacing GP with RF, NGB, or BASS can yield substantial computational savings. Their linear time scaling and negligible GPU footprint make them far more suitable for large-scale or high-throughput BO scenarios, without the prohibitive scaling bottlenecks inherent to GP surrogates.
 
The optimization performance of all surrogates is evaluated with respect to two crucial quantities: compute time and number of acquired samples. Both are important, but their relative weight depends on the nature of the oracle that supplies BO with function evaluations. When the oracle is computational, for instance a lightweight model or a fast equation-based solver, the bottleneck is the BO procedure itself, and compute time becomes the dominant metric. Conversely, when the oracle corresponds to an experiment or an expensive simulation, the true bottleneck is the number of samples, since each evaluation may require hours, days, or even weeks, completely overshadowing the cost of BO.

\begin{figure}[H]
    \centering
    \includegraphics[width=1.1\linewidth]{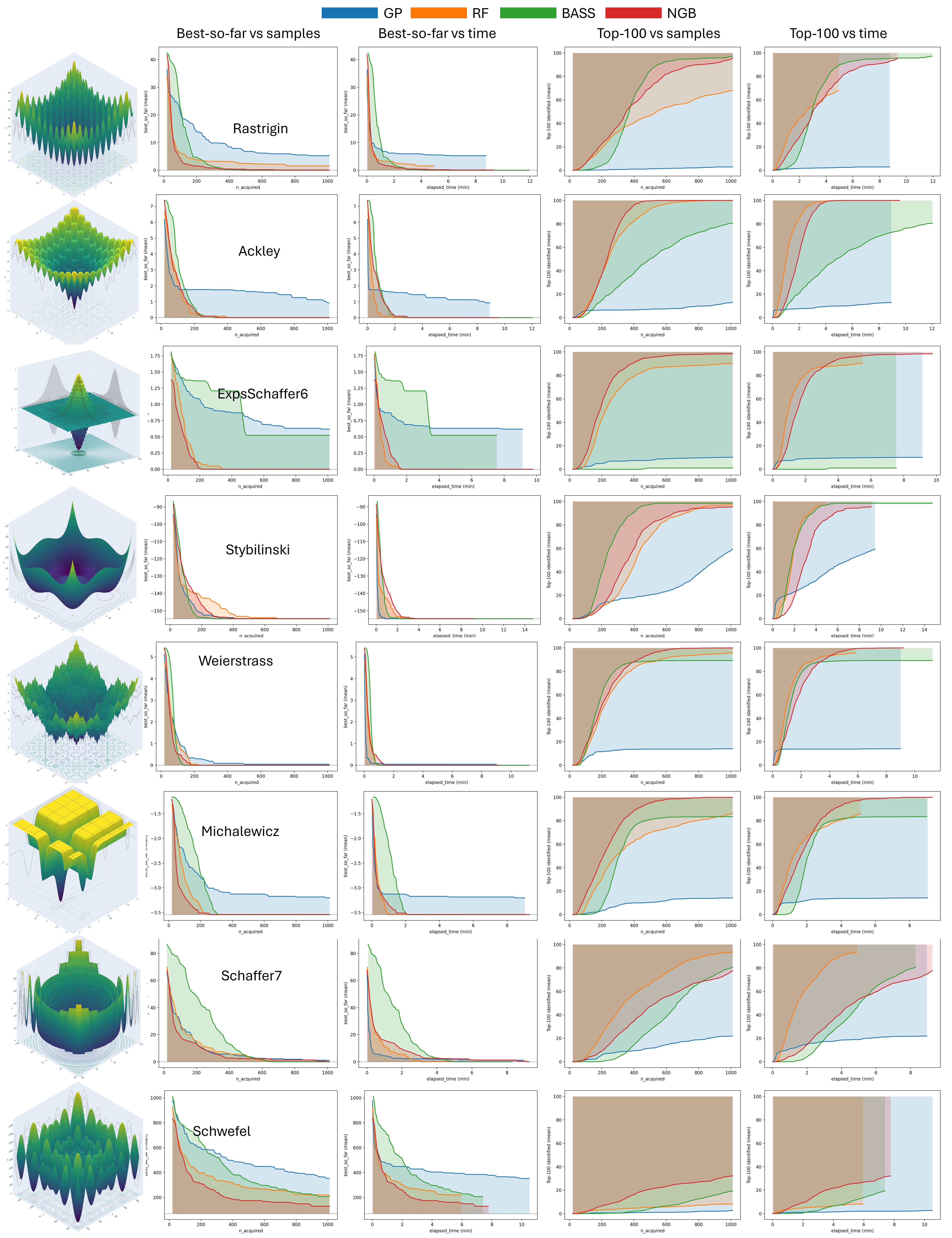}
    \caption{From top to bottom, the eight benchmark functions considered in this study are shown. From left to right, each row displays: (i) a 3D rendering or PCA-based representation of the design space; (ii) the best objective value found as a function of the number of acquired samples; (iii) the best objective value found as a function of compute time; (iv) the number of top–100 global solutions identified as a function of acquired samples; and (v) the same top–100 identification plotted against compute time. Shaded regions denote AUC variability across BO runs, and all curves compare the four surrogate models (GP, RF, BASS, NGBoost) under identical acquisition settings.}
    \label{fig:functions_results}
\end{figure}

To capture performance across both regimes, we report the evolution of (i) the best objective value found and (ii) the identification of the top--100 best solutions, each plotted as a function of compute time and as a function of the number of acquired samples (four plots in total). Because judging which model performs best is not reliable through visual inspection alone, we quantify performance using the area under the curve (AUC); smaller AUC values indicate better overall performance. In the plots, shaded regions illustrate the variability of AUC across BO runs. Figure~\ref{fig:functions_results} summarizes the optimization and compute performance across all benchmark functions.

To obtain a single metric describing BO \emph{compute-time performance}, we multiply the AUC of the ``best objective vs.\ time'' curve with the AUC of the ``top--100 identification vs.\ time'' curve. This product captures both (i) how rapidly a model approaches the global optimum and (ii) how quickly it identifies a large set of high-quality solutions, all measured in wall-clock time. In an analogous fashion, we estimate BO \emph{sample-acquisition performance} by multiplying the AUC of the ``best objective vs.\ samples'' curve with the AUC of the ``top--100 identification vs.\ samples'' curve. \red{All resulting metrics are tabulated in the Supplementary Tables 1--4.}

In terms of BO compute-time performance, RF is the clear winner across all benchmark functions: it reaches the global optimum fastest and identifies a large fraction of the top--100 solutions within minimal compute time. NGB typically ranks second, while GP is last in most cases. Two exceptions occur: on the Styblinski function, BASS ranks second, and on Schaffer7, GP ranks second rather than last. For BO sample-acquisition performance, NGB dominates, ranking first in six out of eight cases. Again, GP is last or second-to-last in the majority of benchmarks.

\begin{comment}
\begin{table}[htbp]
    \centering
    \caption{Ranking of BO surrogate models (1 = best, 4 = worst) in terms of 
    compute-time and sample-acquisition performance across benchmark functions.}
    \label{tab:benchmark_ranks}
    \small
    \begin{tabular}{lcccccccc}
        \toprule
        & \multicolumn{4}{c}{Compute-time performance} 
        & \multicolumn{4}{c}{Sample-acquisition performance} \\
        \cmidrule(lr){2-5} \cmidrule(lr){6-9}
        Function & 1 & 2 & 3 & 4 & 1 & 2 & 3 & 4 \\
        \midrule
        expschaffer6 & RF  & NGB & BASS & GP   & NGB & RF  & GP   & BASS \\
        rastrigin    & RF  & NGB & BASS & GP   & NGB & BASS& RF   & GP   \\
        Michalewicz  & RF  & NGB & BASS & GP   & NGB & RF  & BASS & GP   \\
        Ackley       & RF  & NGB & BASS & GP   & NGB & RF  & BASS & GP   \\
        Schwefel     & RF  & NGB & BASS & GP   & NGB & RF  & BASS & GP   \\
        Stybilinski  & RF  & BASS& GP   & NGB  & BASS& NGB & RF   & GP   \\
        Weierstrass  & RF  & NGB & BASS & GP   & RF  & BASS& NGB  & GP   \\
        Schaffer7    & RF  & GP  & NGB & BASS & RF  & NGB & GP   & BASS \\
        \bottomrule
    \end{tabular}
\end{table}
\end{comment}

Taken together, these results (summarized in Table~\ref{tab:combined_ranks}) reveal a clear and consistent contrast across all benchmark functions: GP-based BO is substantially more expensive in both compute time and GPU memory, yet this overhead does not translate into superior optimization performance. In fact, the opposite trend emerges. RF, BASS, and NGB achieve faster progress toward the optimum \emph{and} identify high-quality solutions with far fewer samples, as reflected in their markedly better compute-time and sample-acquisition AUC metrics. These models therefore deliver both stronger optimization performance and significantly lower computational cost. This establishes a decisive expectation for the real-case datasets: surrogate models with linear scaling and negligible GPU requirements—RF, BASS, and NGB—are likely to provide clear advantages when BO is applied to larger, higher-dimensional, or experimentally constrained problems. We now turn to these real-world datasets to assess whether these trends persist in practical settings.

\subsection{Real-world datasets}
The benchmark functions constitute well-behaved testbeds whose response surfaces are dictated by strict analytical formulas, reflecting idealized patterns inspired by physics, biology, or finance. In contrast, real-world optimization problems rarely exhibit such structure: their landscapes are irregular, their sampling is uneven, and the underlying phenomena are far more intricate. To assess Bayesian Optimization under these realistic conditions, we assembled a suite of nine demanding, domain-spanning datasets. These datasets arise either from (i) in-house, expert-developed modeling pipelines constructed specifically for this work, or (ii) carefully curated and processed literature datasets that required substantial domain knowledge, custom preprocessing, and new Python tooling. Together, they span bio-inspired mechanical engineering, biomechanics, electro-mechanics, drug discovery, materials science, physics, robotics control, and machine learning.

\medskip
\noindent\textbf{In-house, expert-developed datasets}
\begin{itemize}
    \item \textbf{Moiré multilayer materials}: geometric patterns generated by superimposed hexagonal layers with controlled rotations. The objective is to maximize the uniformity of the size distribution of the pores created through the rotation of the stacked layers
    \item \textbf{Bouligand pressure vessel}: tens of thousands of composite laminate designs evaluated using a custom finite-element simulation framework for stress-based performance.
    \item \textbf{Mulit-PZT Semi-active tuned mass damper (SATMD)}: a structural-vibration dataset targeting multi-modal vibration minimization under semi-active control.
    \item \textbf{\emph{C.\ elegans} worm-indentation mechanics}: a mechanics-informed dataset integrating AFM indentation profiles with hyperelastic modeling to quantify drug-induced softening.
    \item \textbf{Lunar Lander control-sequence dataset}: a 50-step discrete action-sequence space defining a black-box control problem through episode-return evaluations.
    \item \textbf{Neural-network hyperparameter tuning}: a wide grid of MLP architectures and training parameters evaluated for predictive performance.
\end{itemize}

\medskip
\noindent\textbf{Curated and processed literature datasets}
\begin{itemize}
    \item \textbf{Ro4 molecular docking}: a curated subset of the ultralarge Ro4 chemical space, processed with RDKit descriptors extraction.
    \item \textbf{COFs dataset}: Monte Carlo simulation data of covalent-organic frameworks for high methane deliverable capacity.
    \item \textbf{QM9 quantum-chemistry dataset}: B3LYP-computed molecular properties for small organic compounds, processed with RDKit descriptors extraction, towards maximizing the LUMO-HOMO gap.
\end{itemize}

\medskip
Information on the construction, processing, and modeling pipelines for all nine datasets is provided in detail in the Methods section.

First of all, it is worth noting that in all real-case datasets the compute-time complexity and GPU-memory scaling as a function of the number of acquired samples follow the same pattern observed for the benchmark functions: Gaussian Processes exhibit a characteristic $N^{3}$ growth in compute time and an approximately linear increase in GPU memory. In contrast, RF, NGBoost, and BASS maintain nearly constant GPU memory usage and display linear compute-time scaling throughout the BO trajectory. \red{All corresponding plots---GPU memory and compute time as a function of sample acquisition---are provided in the ESI (\red{Supplementary Figure~3}).} This again highlights the point raised in the Introduction: GPs impose a substantial computational burden on Bayesian Optimization, making them increasingly strenuous to train as the number of samples grows, whereas the alternative surrogates remain lightweight and scalable.

\begin{figure}[H]
    \centering
    \includegraphics[width=0.9\linewidth]{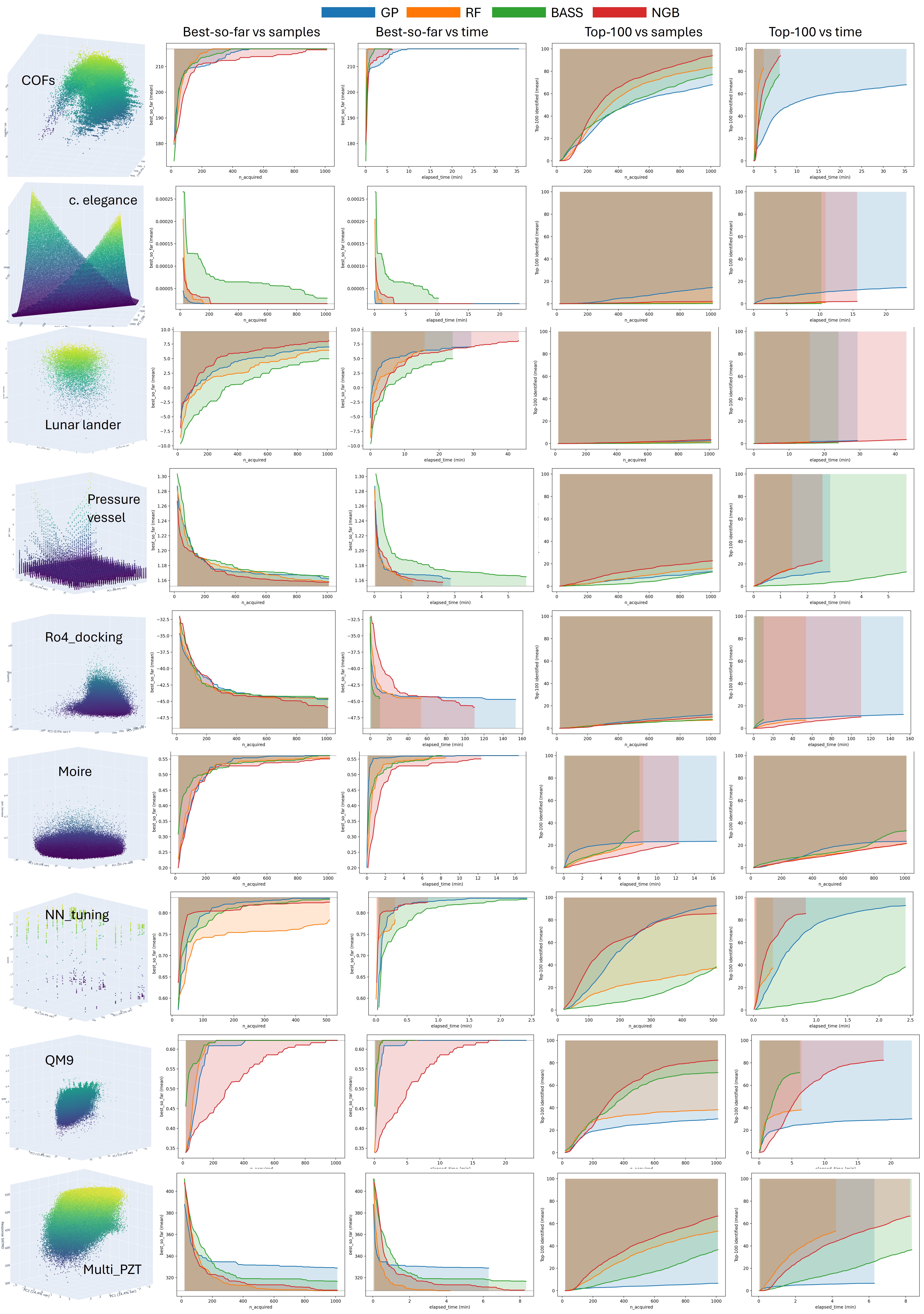}
    \caption{From top to bottom, the eight real-case data considered in this study are shown. From left to right, each row displays: (i) a 3D rendering or PCA-based representation of the design space; (ii) the best objective value found as a function of the number of acquired samples; (iii) the best objective value found as a function of compute time; (iv) the number of top–100 global solutions identified as a function of acquired samples; and (v) the same top–100 identification plotted against compute time. Shaded regions denote AUC variability across BO runs, and all curves compare the four surrogate models (GP, RF, BASS, NGBoost) under identical acquisition settings.}
    \label{fig:realCases_results}
\end{figure}

The analysis that follows examines whether these computational observations also translate into improved BO performance, both in terms of compute time and sample efficiency. As before, the results are presented using the same structure as for the benchmark functions: plots of BO compute-time performance and BO sample-acquisition performance for all four surrogate models, with the area under the curve (AUC) serving as a clear and comparable performance metric across datasets. Figure~\ref{fig:realCases_results} summarizes the optimization and compute performance across all real case datasets.

Across the nine real-case datasets, Gaussian Processes exhibit consistently weak performance in both BO compute-time and sample-acquisition metrics. GP ranks first in compute-time performance only twice (the \textit{Moire} and \textit{worm} datasets) and achieves the top rank in sample-efficiency only once, again in the \textit{worm} case, where the landscape happens to align well with GP smoothness assumptions. In all remaining datasets, GP appears as second-to-last or last, highlighting its limited practical competitiveness once computational cost and scalability are taken into account. By contrast, Random Forest and NGBoost emerge as the strongest overall surrogates: RF attains the best compute-time performance in five out of eight datasets, reflecting its linear scaling and robustness, whereas NGBoost dominates sample-efficiency, ranking first in five datasets. BASS performs well in selected cases but is less consistently dominant across domains. All results are summarized in Table~\ref{tab:combined_ranks}. \red{The values for all the AUC for all cases and surrogates can be found in All resulting metrics are tabulated in the Supplementary Tables 5--8.}

\begin{comment}
\begin{table}[htbp]
    \centering
    \caption{Ranking of BO surrogate models (1 = best, 4 = worst) in terms of 
    compute-time and sample-acquisition performance across real-case datasets.}
    \label{tab:realcase_ranks}
    \small
    \begin{tabular}{lcccccccc}
        \toprule
        & \multicolumn{4}{c}{Compute-time performance} 
        & \multicolumn{4}{c}{Sample-acquisition performance} \\
        \cmidrule(lr){2-5} \cmidrule(lr){6-9}
        Dataset & 1 & 2 & 3 & 4 & 1 & 2 & 3 & 4 \\
        \midrule
        Moire            & GP   & BASS & RF   & NGB  & BASS & GP   & RF   & NGB \\
        pressure\_vessel & RF   & NGB  & GP   & BASS & NGB  & RF   & GP   & BASS \\
        MOFs             & RF   & BASS & NGB  & GP   & RF   & BASS & GP   & NGB \\
        proteins         & BASS & RF   & NGB  & GP   & NGB  & RF   & GP   & BASS \\
        lunar            & RF   & GP   & BASS & NGB  & NGB  & GP   & RF   & BASS \\
        worm             & GP   & RF   & NGB  & BASS & GP   & BASS & RF   & NGB \\
        neural           & RF   & NGB  & GP   & BASS & NGB  & GP   & BASS & RF   \\
        QM9          & BASS   & RF  & GP   & NGB & BASS  & RF   & GP & NGB   \\
        multi\_PZT    & RF   & GP  & NGB   & BASS & NGB  & GP   & RF & BASS   \\
        \bottomrule
    \end{tabular}
\end{table}
\end{comment}

\begin{table}[htbp]
    \centering
    \caption{Ranking of BO surrogate models (1 = best, 4 = worst) in terms of 
    compute-time and sample-acquisition performance across benchmark functions 
    and real-case datasets.}
    \label{tab:combined_ranks}
    \small
    \begin{tabular}{lcccccccc}
        \toprule
        & \multicolumn{4}{c}{Compute-time performance} 
        & \multicolumn{4}{c}{Sample-acquisition performance} \\
        \cmidrule(lr){2-5} \cmidrule(lr){6-9}
        Task / Dataset & 1 & 2 & 3 & 4 & 1 & 2 & 3 & 4 \\
        \midrule
        \multicolumn{9}{c}{\textbf{Benchmark Functions}} \\
        \midrule
        expschaffer6 & RF  & NGB & BASS & GP   & NGB & RF  & GP   & BASS \\
        rastrigin    & RF  & NGB & BASS & GP   & NGB & BASS& RF   & GP   \\
        Michalewicz  & RF  & NGB & BASS & GP   & NGB & RF  & BASS & GP   \\
        Ackley       & RF  & NGB & BASS & GP   & NGB & RF  & BASS & GP   \\
        Schwefel     & RF  & NGB & BASS & GP   & NGB & RF  & BASS & GP   \\
        Stybilinski  & RF  & BASS& GP   & NGB  & BASS& NGB & RF   & GP   \\
        Weierstrass  & RF  & NGB & BASS & GP   & RF  & BASS& NGB  & GP   \\
        Schaffer7    & RF  & GP  & NGB & BASS & RF  & NGB & GP   & BASS \\
        
        \midrule
        \multicolumn{9}{c}{\textbf{Real-Case Datasets}} \\
        \midrule
        Moire            & GP   & BASS & RF   & NGB  & BASS & GP   & RF   & NGB \\
        pressure\_vessel & RF   & NGB  & GP   & BASS & NGB  & RF   & GP   & BASS \\
        MOFs             & RF   & BASS & NGB  & GP   & RF   & BASS & GP   & NGB \\
        Ro4 docking         & BASS & RF   & NGB  & GP   & NGB  & RF   & GP   & BASS \\
        lunar            & RF   & GP   & BASS & NGB  & NGB  & GP   & RF   & BASS \\
        c elegance             & GP   & RF   & NGB  & BASS & GP   & BASS & RF   & NGB \\
        NN tuning           & RF   & NGB  & GP   & BASS & NGB  & GP   & BASS & RF   \\
        QM9              & BASS & RF   & GP   & NGB  & BASS & RF   & GP   & NGB \\
        multi\_PZT       & RF   & GP   & NGB  & BASS & NGB  & GP   & RF   & BASS \\
        
        \bottomrule
     \end{tabular}
\end{table}

\subsection{Building a recommendation system for surrogate models in BO}
\begin{figure}[H]
    \centering
    \includegraphics[width=1.0\linewidth]{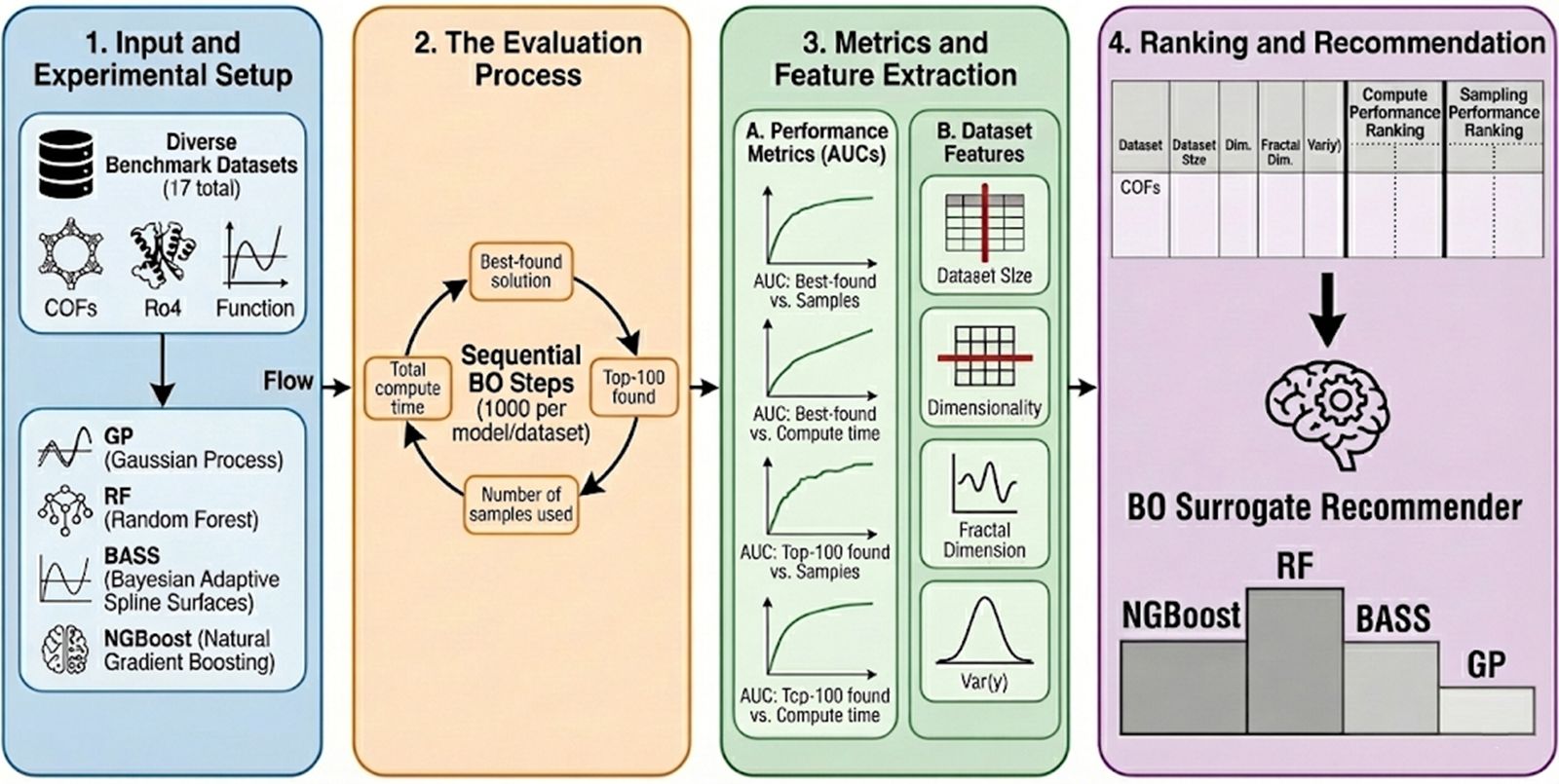}
    \caption{a): our four surrogates evaluation scheme in this work through 8 benchmark functions and 9 real-case datasets; b)Building the classifier that recommends the best surrogate given a dataset} 
    \label{fig:pipeline}
\end{figure}

The left panel of Fig.~\ref{fig:pipeline} summarizes the Bayesian Optimization (BO) evaluation pipeline developed and analyzed throughout this work. Across both benchmark functions and real-world datasets, Gaussian Processes (GPs) consistently emerge as the least practical surrogate: their cubic scaling in training time and steadily increasing memory footprint impose a substantial computational burden, while their sampling efficiency rarely compensates for this cost. In contrast, RF, NGBoost, and BASS achieve comparable or superior optimization performance with markedly lower computational overhead. These results highlight a key insight: no surrogate model is universally optimal, and defaulting to GP-based BO is seldom justified across diverse problem settings.

Motivated by this observation, we move beyond retrospective comparison and introduce a surrogate-model recommendation strategy. The core idea is to use a small set of dataset characteristics that are inexpensive to obtain, yet informative of surrogate performance. Specifically, we consider four features: dataset size (number of samples), dimensionality (number of input features), fractal dimension of the input space, and the variance of the target property. The first three features are extracted directly from the dataset descriptors, while the target variance is estimated from a short initial BO phase, using the first 200 acquisitions with GP as a surrogate.

Based on these features, we train two multi-output classifiers that predict the ranking of the four surrogate models considered in this work (GP, RF, NGBoost, and BASS). One classifier targets compute-time performance, while the other targets sample-acquisition performance. Together, these models act as surrogate recommenders, enabling the selection of an appropriate BO surrogate tailored to the computational or experimental constraints of a given problem. The right panel of Fig.~\ref{fig:pipeline} illustrates the complete workflow used to construct these classifiers, transforming empirical BO benchmarking results into a practical, dataset-aware recommendation system.

To evaluate the performance of the surrogate-ranking classifier, we adopt a 5-fold cross-validation scheme. Each fold is assessed using the Normalized Discounted Cumulative Gain (nDCG)~\cite{scikit-learn}, a standard ranking metric that measures the quality of a predicted ordering relative to an ideal ranking, while placing greater emphasis on correctly identifying higher-ranked (more relevant) items.

The nDCG at position $p$ is defined as
\begin{equation}
    \mathrm{nDCG}_p = \frac{\mathrm{DCG}_p}{\mathrm{iDCG}_p},
    \label{eqn:ndcg}
\end{equation}
where $\mathrm{DCG}_p$ denotes the Discounted Cumulative Gain of the predicted ranking up to position $p$, and $\mathrm{iDCG}_p$ is the corresponding gain for the ideal (perfectly ordered) ranking.

The Discounted Cumulative Gain is given by
\begin{equation}
    \mathrm{DCG}_p = \sum_{i = 1}^{p} \frac{2^{\mathrm{rel}_i} - 1}{\log_2(i+1)},
    \label{eqn:dcg}
\end{equation}
where $\mathrm{rel}_i$ is the relevance score of the item at rank $i$ in the predicted ordering.

The ideal Discounted Cumulative Gain is computed analogously by sorting items according to their true relevance:
\begin{equation}
    \mathrm{iDCG}_p = \sum_{i = 1}^{|REL_p|} \frac{2^{\mathrm{rel}_i} - 1}{\log_2(i+1)}.
    \label{eqn:idcg}
\end{equation}
Here, $p$ denotes the number of items in the ranking, and $|REL_p|$ is the number of relevant items considered.

For each fold, we report:
(i) the mean nDCG across test datasets in that fold, and
(ii) the distribution of per-dataset nDCG values aggregated across all folds.

This analysis is performed separately for rankings derived from sample-efficiency performance and from computational-efficiency (time-based) performance, allowing us to examine whether surrogate selection can be predicted consistently across different notions of frugality.

\begin{figure}[H]
    \centering
    \includegraphics[width=1.0\linewidth]{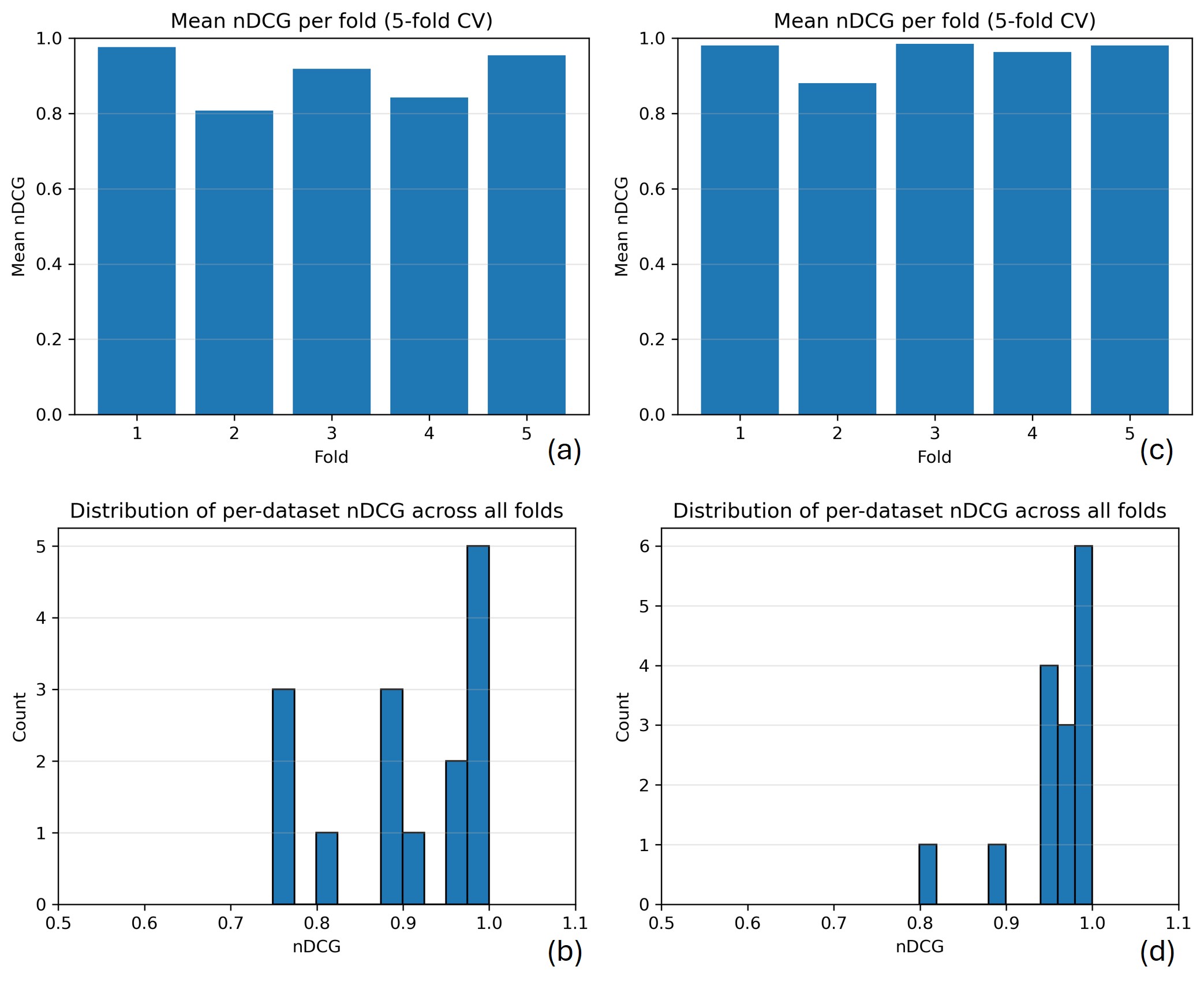}
    \caption{Mean nDCG per fold (top) and per-dataset nDCG distributions aggregated across folds (bottom) are shown for rankings derived from sample-efficiency performance (left) and computational-efficiency performance (right).} 
    \label{fig:classifier_performance}
\end{figure}

In \ref{fig:classifier_performance} we show the performance evaluation of our classifier.

The results demonstrate that the meta-classifier achieves consistently high nDCG values across folds for both performance criteria.

For sample-efficiency rankings (left panels), the mean nDCG per fold remains high, indicating that the predicted surrogate orderings closely match the true rankings derived from BO performance. The per-sample nDCG distribution further shows that most datasets achieve nDCG values near 1, implying that even when the full ranking is not predicted exactly, the classifier typically identifies the most relevant surrogates correctly—particularly at the top of the ranking.

A similar, and in fact slightly stronger, trend is observed for computational-efficiency rankings (right panels). Here, the mean nDCG values are consistently high across all folds, and the per-sample distribution is even more concentrated near nDCG $\approx 1$. This suggests that surrogate performance with respect to computational cost exhibits more regular, learnable structure across datasets, making it particularly amenable to meta-learning.

Overall, these results indicate that the proposed meta-BO framework can reliably infer near-optimal surrogate rankings from simple dataset characteristics, with especially strong performance when targeting computational frugality. The use of nDCG highlights that the classifier is effective not only at exact ranking prediction but, more importantly, at correctly prioritizing the most suitable surrogate models for a given optimization task.

\section{Discussion}
Bayesian Optimization is often presented as a sample-efficient alternative to brute-force exploration, yet in practice its computational overhead is rarely treated as a first-class concern. In this work, we show that this omission is consequential. Across a diverse set of benchmark functions and domain-relevant datasets, surrogate choice fundamentally determines not only optimization performance but also whether BO remains computationally viable as acquisitions accumulate. Our results demonstrate that the long-standing default Gaussian Processes frequently represents the least practical option once wall-clock time and memory usage are explicitly accounted for.

A consistent pattern emerges across all experiments: GP-based BO incurs a rapidly increasing computational and memory burden, while offering little or no compensating advantage in optimization quality. In both benchmark and real-world problems, GP training time grows superlinearly and GPU memory usage accumulates steadily, often becoming the dominant bottleneck even when objective evaluations themselves are inexpensive. By contrast, Random Forests, NGBoost, and BASS exhibit near-linear scaling in compute time and effectively bounded memory usage, allowing BO to proceed smoothly over hundreds to thousands of acquisitions on standard hardware.

Crucially, this improved frugality does not come at the expense of optimization success. On the contrary, the non-GP surrogates consistently match or surpass GP-based BO in identifying optimal or near-optimal solutions, both in terms of best-found values and recovery of top-performing candidates. These findings challenge the widespread assumption that GP smoothness and calibrated uncertainty inherently translate into superior BO performance. Instead, they suggest that, for many realistic design spaces, the computational structure of the surrogate model matters at least as much as its theoretical optimality guarantees.

While GP-based BO performs poorly on average, no single alternative surrogate dominates across all datasets and performance regimes. Random Forests excel in compute-time–limited settings, while NGBoost often achieves superior sample efficiency; BASS provides strong performance in selected domains with complex, nonstationary structure. This heterogeneity reinforces a central conclusion of this study: there is no universally optimal surrogate model for Bayesian Optimization.

This observation has important practical implications. In experimental science, where sample acquisition is expensive, optimizing for sample efficiency may be paramount. In simulation-driven or data-rich settings, compute time and memory footprint may dominate instead. Treating BO as a one-size-fits-all procedure, by defaulting to GP surrogates, obscures these trade-offs and can lead to unnecessarily costly optimization pipelines. Our results argue for a shift in perspective: surrogate selection should be viewed as a dataset-dependent decision rather than a fixed design choice.

Motivated by this insight, we extend FruBO beyond retrospective benchmarking and introduce a data-driven surrogate recommendation framework. Using a small set of inexpensive dataset characteristics (dataset size, dimensionality, fractal dimension, and a lightweight estimate of target variability) we show that it is possible to predict surrogate rankings with high fidelity. The strong nDCG scores achieved by our multi-output classifiers indicate that surrogate performance exhibits learnable structure across datasets, particularly when computational efficiency is the target.

This recommendation layer converts empirical benchmarking into a practical decision tool. Instead of relying on costly pilot studies across multiple surrogates, FruBO enables surrogate selection \emph{a priori}, aligned with either compute-time–limited or sample-limited regimes. In doing so, it moves beyond comparison and embeds computational frugality directly into the BO workflow.

The implications extend beyond Bayesian Optimization. As machine learning becomes integral to scientific discovery, computational cost and energy usage can no longer be treated as secondary concerns. Even modest workloads, when repeated across models and optimization loops, accumulate substantial overhead. Our results indicate that compute-aware evaluation protocols can deliver immediate efficiency gains without compromising optimization quality.

By reframing BO evaluation along two axes (optimization quality and computational frugality) this work provides a template for future methodological studies in scientific machine learning. We anticipate that similar paired evaluations will be increasingly necessary as the community confronts tighter compute budgets, growing environmental concerns, and the need to democratize advanced optimization tools beyond specialized HPC environments.

Several limitations of the present study point to directions for future work. First, while we consider four representative surrogates, the landscape of scalable probabilistic models continues to expand, including sparse GP variants, neural surrogates, and hybrid ensembles. Extending the FruBO framework to include these models would further enrich the recommendation space. Second, our dataset-characteristic features are intentionally simple; incorporating richer descriptors of landscape structure or noise could further improve surrogate selection, particularly in small-data regimes. Finally, while our recommendation system focuses on surrogate choice, future extensions could jointly recommend acquisition functions or dynamically adapt surrogate models during the BO loop.

Despite these limitations, the central conclusion is clear: Bayesian Optimization need not be computationally heavy to be effective. By abandoning the default reliance on Gaussian Processes and embracing scalable alternatives guided by data-aware recommendations, BO can become a genuinely frugal and widely applicable tool for scientific discovery.

\section{Methods}\label{sec:methods}
FruBO is implemented on top of the BoTorch framework~\cite{botorch}, which provides the core Bayesian Optimization abstractions and Gaussian-process baselines. FruBO extends this foundation by systematically integrating and benchmarking alternative, scalable surrogate models beyond standard GP-based BO, including Random Forests, NGBoost, and Bayesian Adaptive Spline Surfaces (BASS). The framework is designed to be modular, allowing users to easily incorporate additional surrogate models and evaluate them under unified acquisition, evaluation, and logging protocols. The full implementation is publicly available; see the Code Availability section.

\subsection{Surrogate models}
\subsection*{Surrogate Models}

We evaluate four surrogate models within our Bayesian Optimization framework, chosen to balance predictive performance and computational efficiency.

\paragraph{Gaussian Processes (SingleTaskGP).}
We adopt the standard exact GP implementation in BoTorch \cite{botorch}, using automatic relevance determination (ARD) kernels and outcome standardization. GPs remain the canonical choice in BO due to their principled uncertainty quantification, though their training cost scales cubically with the number of observations and memory usage increases linearly.

\paragraph{Random Forest (RF) surrogate.}
\emph{Random Forests} (RFs) are nonparametric ensemble models that combine multiple regression trees to approximate complex, nonlinear response surfaces. We include RF as a frugal and well-established baseline surrogate, widely adopted in model-based optimization and algorithm configuration~\cite{Hutter2011}. RFs naturally handle nonlinear, nonstationary, and mixed-type inputs without requiring kernel definitions or feature scaling, and their training cost scales nearly linearly with the number of samples and trees, making them particularly suitable for iterative Bayesian Optimization loops. Compared to Gaussian Processes, RFs avoid the $\mathcal{O}(n^3)$ covariance inversion bottleneck while providing variance estimates from the ensemble dispersion, albeit with coarser calibration. In contrast to more sophisticated surrogates such as BASS and NGBoost, RFs are computationally lighter, require minimal hyperparameter tuning, and are robust to outliers and discontinuities. 

We use the standard \texttt{RandomForestRegressor} implementation from \texttt{scikit-learn} with default parameters. Uncertainty is estimated from the variance across individual trees. While RF is less theoretically grounded in its uncertainty quantification compared to GPs, it remains computationally efficient, robust to noise, and scales well to large datasets. Their interpretability, mature implementations, and strong empirical record in sequential model-based optimization (SMBO) frameworks (e.g., SMAC~\cite{Hutter2011}), makes RF a good alternative to GP as scalable BO surrogates \cite{Styrud2024}.

\paragraph{Bayesian Adaptive Spline Surfaces (BASS).}
BASS \cite{Francom2020} is a nonparametric Bayesian regression model based on adaptive spline bases with posterior inference via reversible-jump MCMC sampling. It extends the Bayesian Multivariate Adaptive Regression Splines (BMARS) framework of~\citet{Denison1998} by supporting both continuous and categorical inputs, scalar or functional outputs, and by incorporating more flexible priors and Reversible Jump Markov Chain Monte Carlo (RJMCMC) strategies for exploring model space. BASS is also equipped with analytical Sobol decompositions, making it directly useful for sensitivity analysis. These enhancements make BASS more versatile and scalable than BMARS for high-dimensional or mixed-type design spaces.

We include BASS among our surrogate models because it provides a computationally frugal, nonstationary-capable, and uncertainty-aware alternative to Gaussian Processes. Its adaptive spline basis allows the model to vary smoothness across the input space without assuming stationarity, capturing heterogeneous regions that often appear in physical and benchmark functions. The RJMCMC inference mechanism automatically adjusts model complexity by adding or pruning basis functions, while shrinkage priors control overfitting as the BO loop progresses. Compared to BMARS, BASS offers improved scalability, parallelization options, and more efficient posterior exploration through parallel tempering. It natively handles categorical predictors and functional responses, and its full Bayesian formulation yields structured uncertainty estimates readily usable by acquisition functions. Together, these features make BASS particularly suited to frugality-oriented Bayesian Optimization, where flexibility, interpretability, and efficient uncertainty quantification are required under limited evaluation budgets. Across domains, BASS has demonstrated state-of-the-art predictive accuracy and efficiency. In building-energy modeling, multi-output BASS achieves CV(RMSE) < 0.005 with $R^2 \approx 1$, outperforming deep-learning baselines while preserving correlations among multiple outputs, and its analytical Sobol’ implementation reproduces sampling-based rankings with 46$\times$--148$\times$ faster sensitivity computation~\cite{Li2022, Zhang2024}. In materials and nuclear-forensics applications, BASS enables accurate inverse prediction of plutonium processing conditions---on average within one standard deviation of the true experimental settings~\cite{Ausdemore2022}---and shows top performance in predicting detonation metrics such as detonation velocity ($V_{\mathrm{det}}$) and detonation pressure ($P_{\mathrm{det}}$) among several machine-learning models~\cite{Davis2024}. These results confirm BASS as a robust and high-performing surrogate for complex, high-dimensional, and mixed-type problems. However, to our knowledge, it has not yet been implemented or evaluated within a full Bayesian Optimization framework, which motivates its integration and benchmarking in this work.

BASS was originally developed as an R package \cite{Francom2020} (see manual at \url{https://cran.r-project.org/web/packages/BASS/BASS.pdf}) but has recently been ported to Python as \texttt{pyBASS} (\url{https://github.com/lanl/pyBASS/tree/main}), which we employ in this work. We follow settings close to the defaults, but with an even more frugal choice of burn-in: whereas the default is $nmcmc=10000$ and $nburn=9000$ \cite{Francom2020}, we use $nmcmc=10000$ and $nburn=9900$. This reduces the number of effective posterior samples, but substantially lowers computational cost when used inside a repetitive BO pipeline. In this sense, our configuration prioritizes frugality over extracting the very highest predictive performance.  
\begin{verbatim}
nmcmc     = 10000   % total MCMC iterations
nburn     = 9900    % burn-in iterations (default 9000)
thin      = 1
w1, w2    = 5.0, 5.0
maxInt    = 2
maxBasis  = 1000
g1, g2    = 1.0, 1.0
s2_lower  = 0.0
h1, h2    = 10.0, 10.0
a_tau     = 0.5
b_tau     = 1.0
verbose   = True
\end{verbatim}

\paragraph{Natural Gradient Boosting (NGBoost).}
\emph{NGBoost} is a modular probabilistic boosting framework that uses the natural gradient to learn full predictive distributions with stable training dynamics and strong scalability, offering both accurate point predictions and well-calibrated uncertainty estimates~\cite{Duan2020}. Empirically, in low-data chemical design and Bayesian Optimization (BO) benchmarks, NGBoost achieved among the top fractions of hits (e.g., Delaney dataset $0.959$, Freesolv $0.953$) and exhibited strong early-stage optimization and calibration performance, often matching Gaussian Processes around $\sim$100 samples~\cite{Tom2023}. In engineering optimization, a BO--NGBoost surrogate for tunnel deformation prediction attained a test $R^2 \approx 0.92$ and outperformed Random Forest, XGBoost, LightGBM, GRU, and LSTM models, enabling multi-objective improvements up to $\sim$56--62\%~\cite{Chen2025}. Taken together, these results confirm NGBoost as a robust and high-performing surrogate capable of providing reliable probabilistic predictions under limited data, motivating its inclusion and systematic benchmarking within our Bayesian Optimization framework.
 
In principle, NGBoost defaults to a relatively conservative configuration:
\begin{quote}
\begin{verbatim}
n_estimators=500, learning_rate=0.01, minibatch_frac=1.0,
col_sample=1.0, verbose=True, verbose_eval=100, tol=1e-4,
random_state=None, validation_fraction=0.1
\end{verbatim}
\end{quote}
However, in practice, Duan et al.\ (2020) also demonstrate that a larger learning rate ($0.1$) is appropriate for large datasets. We therefore follow this latter choice and at the same time reduce the number of estimators to $100$. This configuration is more computationally economical, making it suitable for iterative use within BO, where many surrogate fits are required. While this choice may sacrifice some predictive accuracy compared to the full defaults, it aligns with our emphasis on frugality and adaptability to medium-scale compute resources.
\begin{verbatim}
n_estimators          = 100   % reduced from default 500
learning_rate         = 0.1   % increased from default 0.01
random_state          = 42
natural_gradient      = True
score                 = "LogScore"
verbose               = False
early_stopping_rounds = None
sigma_floor           = 1e-2
sigma_cap             = 1e2
\end{verbatim}

\medskip
Together, these four surrogates allow us to benchmark classical GP-based BO against alternative ensemble and boosting approaches that bypass the cubic scaling bottleneck of GPs while offering competitive or superior sample efficiency. Importantly, for both BASS and NGBoost we deliberately adopt computationally frugal settings, as our goal is not to maximize surrogate accuracy in isolation, but to evaluate their effectiveness within the demanding, repetitive context of Bayesian Optimization.

\subsection{Datasets}
\subsubsection{COFs (methane uptake and deliverable capacity).}

\datasetstats{69{,}839}{20}{deliverable capacity}{maximize}

Mercado et al.~\cite{Mercado2018} reported $\sim$70{,}000 COFs with Monte Carlo simulations for $CH_4$ uptake and deliverable capacity. Here we target deliverable capacity as separate objectives within our BO framework.

\subsubsection{QM9 Dataset and Molecular Descriptor Extraction.}

\datasetstats{133{,}885}{46}{HOMO-LUMO gap}{maximize}

We employed the QM9 dataset introduced by~\citet{Ramakrishnan2014}, which provides B3LYP/6-31G(2df,p)-level 
quantum-chemical properties for 133{,}885 stable organic molecules composed of 
C, H, N, O and F. The dataset is distributed as extended \texttt{.xyz} files, where 
each entry contains not only the atomic coordinates but also all computed scalar 
properties (including orbital energies, polarizabilities, heat capacities, 
thermochemical quantities, and the HOMO--LUMO gap). Crucially for our workflow, 
each file additionally stores two SMILES representations (the original GDB-17 
SMILES and a geometry-relaxed SMILES), enabling deterministic reconstruction of 
each molecule within RDKit for descriptor calculation.

We did not use 
the full RDKit descriptor library but 
instead retained a subset of 43 physico-chemical descriptors appropriate 
for QM9. These include global size and composition features (e.g.\ \texttt{MolWt}, 
\texttt{ExactMolWt}, \texttt{NumValenceElectrons}), topological and complexity 
indices (\texttt{Chi$n$}, \texttt{Kappa$n$}, \texttt{BalabanJ}, \texttt{Ipc}), 
polarity-related descriptors (\texttt{TPSA}, \texttt{MolLogP}, 
EState indices), and ring- and functionality-count descriptors. 
This reduced set avoids redundancy, excludes descriptors that are 
identically zero within the restricted chemical space of QM9, and captures the 
major structural factors that influence frontier-orbital energetics. 
For the present analysis, the target property was the HOMO--LUMO gap 
(\texttt{gap}) provided directly in the QM9 files.

\subsubsection{Ro4 docking dataset.}

\datasetstats{397{,}104}{210}{binding energy}{minimize}

We further incorporated a large-scale dataset originating from the recent work of Lüttens \textit{et al.}~\cite{Luttens2025}, who combined molecular docking with machine learning to accelerate ultralarge virtual screening campaigns. The dataset, deposited on Zenodo (record \href{https://doi.org/10.5281/zenodo.7953917}{10.5281/zenodo.7953917}), contains docking scores for a multi-billion “rule-of-four” (Ro4) chemical subspace derived from the Enamine REAL library. Ro4 molecules satisfy the property constraints of molecular weight $<$400~Da and cLogP $<$4, and were docked against a set of therapeutically relevant protein targets. The primary target considered here is the A$_{2A}$ adenosine receptor (A2AR), for which the dataset provides per-compound docking energies (kcal/mol). We adopt these docking scores as the optimization objective, treating them as a black-box property to be maximized in the BO loop.

Because the raw dataset comprises billions of molecules, we reduced its size for tractability. Specifically, we first sampled approximately 5\% of the available entries, and then retained 60\% of this subset after filtering out compounds with docking energies worse than 5000~kcal/mol. This yielded a final working dataset of 397{,}104 molecules. Each compound was represented by the full set of 208 RDKit molecular descriptors (physicochemical, topological, and electronic features), resulting in a 209-dimensional dataset once the docking score was included as the target property. This preprocessing pipeline produced a compact yet chemically diverse benchmark that preserves the challenging distribution of docking scores while remaining computationally manageable for surrogate modeling within FruBO.

\subsubsection{Lunar Lander Benchmark dataset}

\datasetstats{500{,}000}{20}{episode return}{maximize}

\paragraph{Dataset motivation and scope}
To evaluate our sampling and optimization strategies on a real control problem, we implemented a discrete action--sequence benchmark based on the \texttt{LunarLander-v3} environment from Gymnasium (OpenAI Gym). Our formulation is inspired by the active optimization setting introduced ~\citet{Wei2024DANTE}, where the classical reinforcement-learning control task is reformulated as a black-box design-space optimization problem. Instead of learning a policy, the objective is to discover the best open-loop sequence of discrete actions that maximizes the total episode reward.

\paragraph{Design space}
The environment exposes a discrete action space
\[
\mathcal{A} = \{0,1,2,3\} = \{\text{no-op},\,\text{left engine},\,\text{main engine},\,\text{right engine}\},
\]
and each candidate solution is represented as a fixed-length action vector.
While Wei~\textit{et al.} employ a horizon of 100 steps (resulting in a $100$-dimensional design space with $4^{100}$ possible action sequences), we use a compressed horizon of \textbf{50 steps} ($K = 50$), which preserves the structure of the optimization problem while reducing temporal complexity and computational cost. Each candidate action sequence
\[
a = (a_{1}, a_{2}, \ldots, a_{50}), \quad a_{t} \in \mathcal{A},
\]
is executed deterministically from a fixed initial state using a controlled random seed. During execution, the environment applies its native reward shaping (distance-to-target minimization, leg-contact bonuses, fuel usage and crash penalties), and the cumulative sum of these rewards is recorded as a scalar \texttt{episode\_return}. For dataset construction, we uniformly sample sequences from $\mathcal{A}^{50}$ and evaluate them in parallel, producing rows of the form:
\[
(a_{1}, a_{2}, \ldots, a_{50}, \texttt{episode\_return}).
\]
This results in a discrete, high-dimensional, black-box design space analogous to the formulation of Wei~\textit{et al.}, but with a reduced horizon ($50$ vs.\ $100$ actions) enabling more efficient benchmarking of frugal optimization methods.

\paragraph{Objective for BO: Reward function.}
In the \texttt{LunarLander-v3} environment, the objective of each
action sequence is to maximize the cumulative return obtained from a shaped
reward function that guides the lander toward a soft, upright touchdown.
Let the lander state at time $t$ be
\[
(x_t,y_t,\dot{x}_t,\dot{y}_t,\theta_t,\dot{\theta}_t,c_{1,t},c_{2,t}),
\]
where $(x_t,y_t)$ is the position relative to the landing pad, 
$(\dot{x}_t,\dot{y}_t)$ are translational velocities,
$\theta_t$ and $\dot{\theta}_t$ the orientation and angular velocity,
and $c_{1,t},c_{2,t}\in\{0,1\}$ indicate left- and right-leg ground contact.
The instantaneous reward used by \texttt{LunarLander-v3} is
\begin{equation}
\label{eq:lunar_reward_step}
\begin{aligned}
r_t = &~
-100\,\sqrt{x_t^2 + y_t^2}
-100\,\sqrt{\dot{x}_t^2 + \dot{y}_t^2}
-100\,|\theta_t|
-10\,|\dot{\theta}_t|  \\
&\quad + 10\,c_{1,t} + 10\,c_{2,t}
+ R_{\text{engine}}(a_t),
\end{aligned}
\end{equation}
where the engine-use penalty is
\[
R_{\text{engine}}(a_t) = 
\begin{cases}
-0.30, & a_t = \text{main engine},\\[3pt]
-0.03, & a_t \in \{\text{left engine},\,\text{right engine}\},\\[3pt]
\;\;0, & a_t = \text{no-op}.
\end{cases}
\]
A terminal bonus or penalty is applied upon episode termination:
\[
r_{\text{terminal}} =
\begin{cases}
+100, & \text{successful landing with both legs down},\\[2pt]
-100, & \text{crash or unstable contact}.
\end{cases}
\]
The overall objective returned for an action sequence
$a=(a_1,\ldots,a_K)$ is therefore
\begin{equation}
\label{eq:lunar_total_return}
\texttt{episode\_return}(a)
=
\sum_{t=1}^{T} r_t \;+\; r_{\text{terminal}},
\end{equation}
where $T\leq K$ is the first time step at which the episode terminates
(success, crash, or timeout).
This scalar quantity is used as the target value in our Bayesian Optimization
experiments.

\begin{figure}[H]
    \centering
    \includegraphics[width=1.0\linewidth]{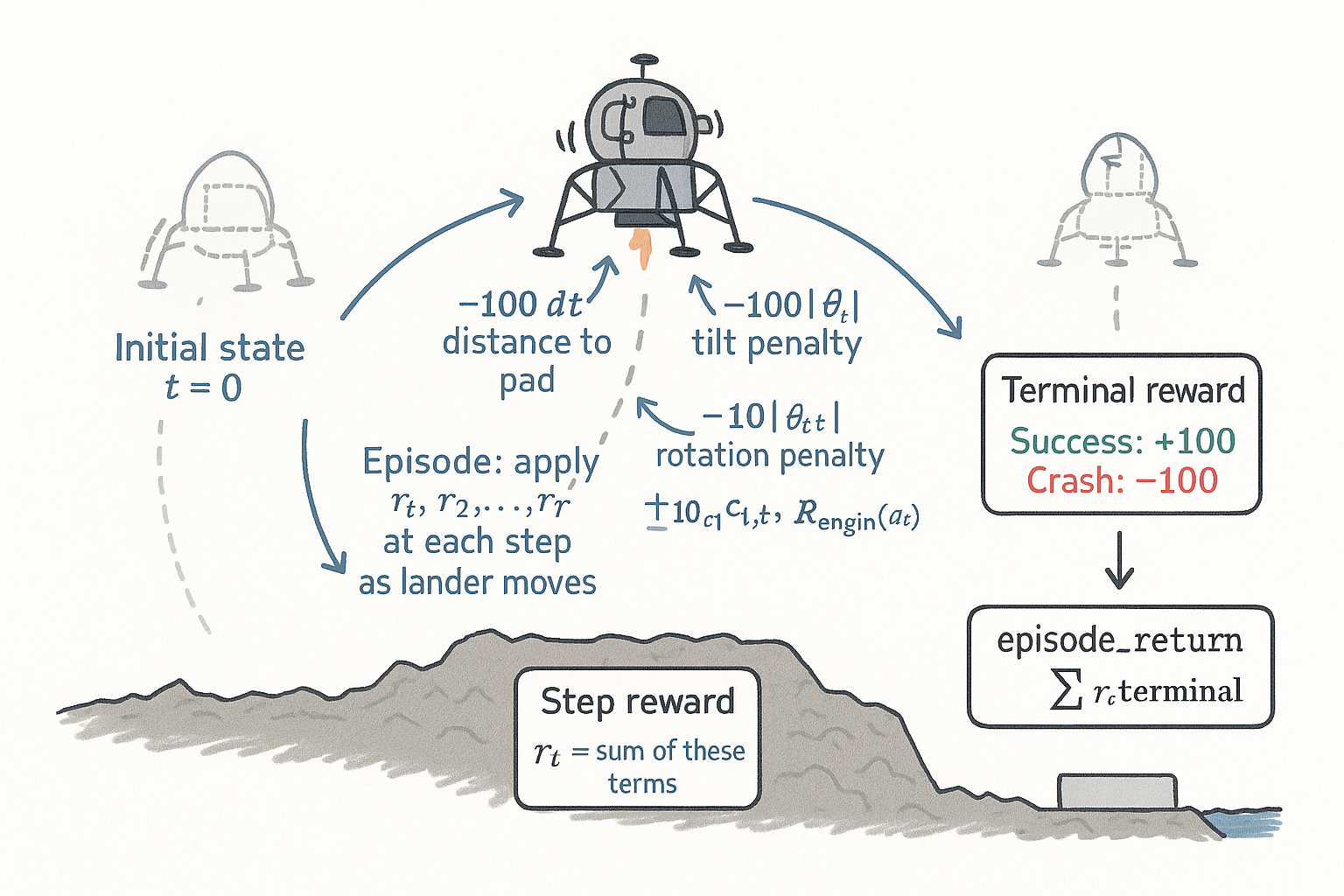}
    \caption{Schematic illustration of the LunarLander reward construction. An open-loop action sequence generates a trajectory from an initial to a terminal
state. At each time step, state-dependent shaping terms and action-dependent engine-use penalties contribute to the step reward $r_t$, while a terminal reward is added upon landing, crash, or timeout to yield the final episode return.} 
    \label{fig:lunar}
\end{figure}

Figure~\ref{fig:lunar} schematically summarizes the reward construction used in the LunarLander benchmark. Starting from a fixed initial state, a candidate open-loop action sequence induces a trajectory of lander
states. At each time step, the instantaneous reward $r_t$ is formed by combining state-dependent shaping terms—penalizing distance to the landing pad, velocity,
tilt, and angular rate—with action-dependent engine-use penalties. Upon landing, crash, or timeout, a terminal reward $r_{\text{terminal}}$ is applied. The final
objective value, \texttt{episode\_return}, is obtained by summing all step rewards and the terminal contribution, as defined in Eqs.~(\ref{eq:lunar_reward_step})–(\ref{eq:lunar_total_return}).

\subsubsection{MLP Hyperparameter Grid Dataset}

\datasetstats{8{,}640}{18}{$R^2$}{maximize}

To construct a supervised dataset for neural-network hyperparameter tuning, we systematically evaluated a feed-forward multilayer perceptron (MLP) on the \textit{Lunar\_landing} regression task. We defined a discrete grid over architectural and training hyperparameters as follows: hidden-layer architectures with depth $\in \{2,3,4,5,6\}$ and uniform width per layer $\in \{8,16,64,128\}$ (20 architectures in total); activation function $\in \{\texttt{relu},\ \texttt{tanh},\ \texttt{logistic}\}$; solver fixed to \texttt{adam}; L2 regularization strength $\alpha \in \{10^{-6},10^{-5},10^{-4},10^{-3},10^{-2},10^{-1}\}$; mini-batch size $\in \{32,64,128,256\}$; and initial learning rate $\in \{10^{-6},10^{-5},10^{-4},10^{-3},10^{-2},10^{-1}\}$. Early-stopping controls were fixed to \texttt{max\_iter}=50, \texttt{tol}=5$\times 10^{-2}$, and \texttt{n\_iter\_no\_change}=10. This grid results in \textbf{8640 unique hyperparameter configurations}, each of which was trained using standardized inputs (\texttt{StandardScaler}) and evaluated using the coefficient of determination ($R^2$), computed via two ShuffleSplit folds with 20\% training and 80\% validation data under a fixed random seed. The grid evaluation was executed in parallel using \texttt{joblib} with 15 worker processes and one BLAS thread per process to avoid oversubscription. All evaluated configurations, together with their validation $R^2$ and training time, were stored in a comprehensive CSV file, from which we derived a compact ``slim'' dataset (\texttt{nn\_grid\_lunar\_score.csv}) containing only the hyperparameters (including numeric summaries of architecture depth and width, and one-hot encodings of activation and solver choices) and the corresponding validation score, with duplicate settings removed.

\subsubsection{Aldicarb-Induced Softening: Mechanics-Informed Inverse-Design Dataset}

\datasetstats{579{,}681}{3}{mse}{minimize}

\textbf{\textit{Dataset motivation and scope}}

\begin{figure}[H]
    \centering
    \includegraphics[width=1.0\linewidth]{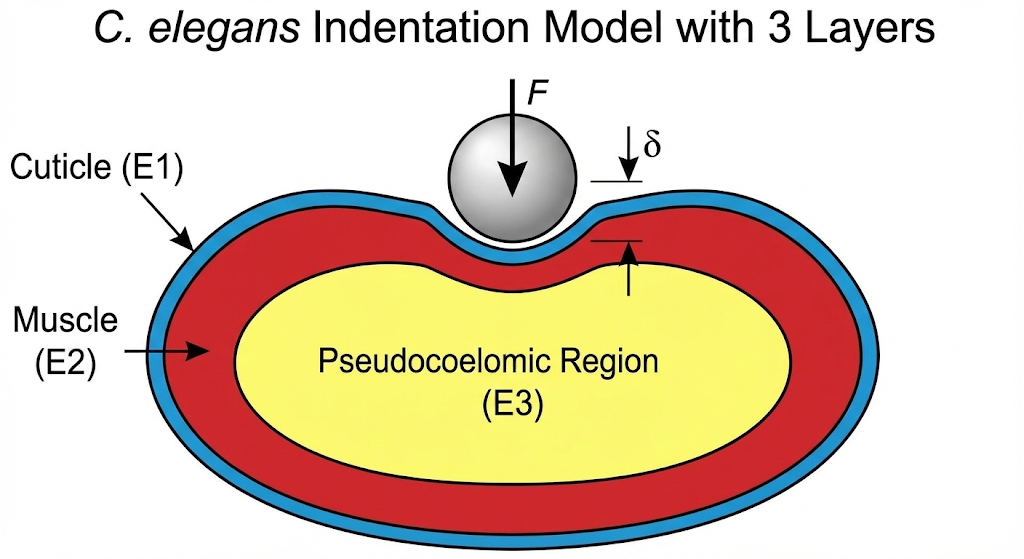}
    \caption{Schematic cross-section of the three-layer indentation model of \emph{C.~elegans} used in this work. An AFM indenter applies a normal force $F$
and induces a displacement $\delta$ on the worm body. The organism is modeled as a multilayered structure comprising an outer cuticle layer ($E_1$), an intermediate muscle layer ($E_2$), and an inner pseudocoelomic region ($E_3$). These effective Young’s moduli parameterize the tissue-scale mechanical response and define the inverse-design space explored via Bayesian Optimization.} 
    \label{fig:worm}
\end{figure}

Neuromuscular agents such as aldicarb have long served as model compounds in \emph{C.~elegans} pharmacological assays, traditionally probed through behavioral readouts such as motility or paralysis~\cite{Elmi2017}. Recent empirical and in-silico work, however, has revealed that aldicarb produces a pronounced mechanical phenotype: instead of hyper-contraction, the dominant effect is a \emph{bulk softening} of the organism arising primarily from depressurization of
the pseudocoelom~\cite{Essmann2024}. This process leads to a stiffness reduction of approximately 66\% relative to BDM-treated controls and correlates with volumetric shrinkage and decreased internal hydrostatic pressure. As a result,
aldicarb acts as a multi-tissue modulator of organismal biomechanics rather than a purely neuromuscular agent. Building upon this characterization, we constructed a mechanics-informed dataset representing the limiting mechanical regime of aldicarb-induced softening. The dataset is designed specifically for benchmarking surrogate-model-based Bayesian Optimization methods on a large-scale, nonlinear inverse problem. The underlying experimental data consist of atomic force microscopy (AFM) force--displacement ($F$--$\delta$) curves from worms exposed to aldicarb in the high-softening regime. These measurements provide the ground-truth mechanical response against which candidate material parameter sets are evaluated. Following the multilayered anatomical structure of \emph{C.~elegans}, three
effective Young’s moduli are considered:
\[
E_1 \text{ (cuticle)},\qquad
E_2 \text{ (muscle layer)},\qquad
E_3 \text{ (pseudocoelomic region)}.
\]
These parameters define the elastic state of the organism in a simplified,
axisymmetric indentation model, consistent with Neo-Hookean hyperelasticity, incompressibility ($\nu = 0.5$), and previously validated finite-element representations~\cite{Rekatsinas2024}. Figure~\ref{fig:worm} illustrates the three-layer mechanical model of \emph{C.~elegans} used to relate tissue-level elastic properties to AFM indentation force--displacement responses.

\textbf{\textit{Design space}}

To build a comprehensive benchmark for BO, we enumerated a full Cartesian grid
in the space of layer-wise elasticities using experimentally informed bounds:
\[
E_1 \in [30, 250], \qquad
E_2 \in [300, 3000], \qquad
E_3 \in [30, 500].
\]
A uniform discretization step of $\Delta E = 5$\,kPa was applied to all axes.
After filtering non-physical combinations, the resulting design space contained \textbf{579,681 unique elasticity triplets}. Each triplet $(E_1,E_2,E_3)$ represents a candidate mechanical phenotype of the worm.

For each configuration, we use a nonlinear multilayer stiffness model as presented by \citet{Rekatsinas2024}. This model computes the
predicted indentation force profile $F_{\mathrm{pred}}(\delta)$ over the
experimentally measured displacements. The model combines:
\begin{enumerate}
    \item Hertzian and post-Hertzian indentation regimes with
          displacement-dependent power-law exponents;
    \item beam-theoretic bending stiffnesses of the cuticle, muscle, and gut
          layers through geometry-specific moments of inertia;
    \item hyperelastic correction factors capturing nonlinear deformation at
          large indentation depths;
    \item a series coupling between indenter stiffness and tissue stiffness.
\end{enumerate}
This generates a deterministic and physically well-grounded forward map
\[
(E_1, E_2, E_3) \ \longmapsto\ F_{\mathrm{pred}}(\delta).
\]

\paragraph{Objective for BO: Force-Matching Error.}
To quantify how well each candidate parameter set reproduces the experimental
indentation curve, we compute the mean squared error:
\[
\mathrm{MSE}(E_1,E_2,E_3)
= \frac{1}{N} \sum_{i=1}^N
\Bigl(F_{\mathrm{pred}}(\delta_i;E_1,E_2,E_3)
      - F_{\mathrm{true}}(\delta_i)\Bigr)^2.
\]
Each row of the dataset thus contains three input features $(E_1,E_2,E_3)$ and a
single scalar regression target (MSE), forming a high-dimensional, non-convex,
anisotropic inverse-design landscape ideally suited for testing BO
surrogates and acquisition strategies.

\subsubsection{High-pressure vessel}

\datasetstats{52{,}272}{5}{scalarized objective $J$}{minimize}

\textbf{\textit{Dataset motivation and scope}}

Although Bouligand (helicoidal) laminates are widely recognized for their ability to
twist cracks, diffuse damage, and suppress delamination, no consolidated,
machine-learning-ready dataset exists for pressure-retaining CFRP shells.
Existing studies highlight the promise of helicoidal architectures---including graded,
hemi-symmetric, and discontinuous variants---for improving damage tolerance
\cite{Loutas2025}, yet raw data are sparse or unavailable.
To support surrogate-model comparison under realistic mechanical conditions, we
constructed a large, structured design--response dataset tailored to Bayesian
Optimization (BO) for architected composite materials.

\textbf{\textit{Design space}}

The Bouligand generator spans five discrete, physically interpretable parameters:
starting angle ($0{:}5{:}175^\circ$), inter-ply pitch ($5{:}5{:}55^\circ$),
ply count ($8{:}1{:}40$), symmetry flag $\{0,1\}$, and ply-thickness mode $\{1,2\}$.
The full enumeration ($36\times11\times33\times2\times2$) yields
$52{,}272$ unique laminate layups.

\begin{figure}[H]
    \centering
    \includegraphics[width=1.0\linewidth]{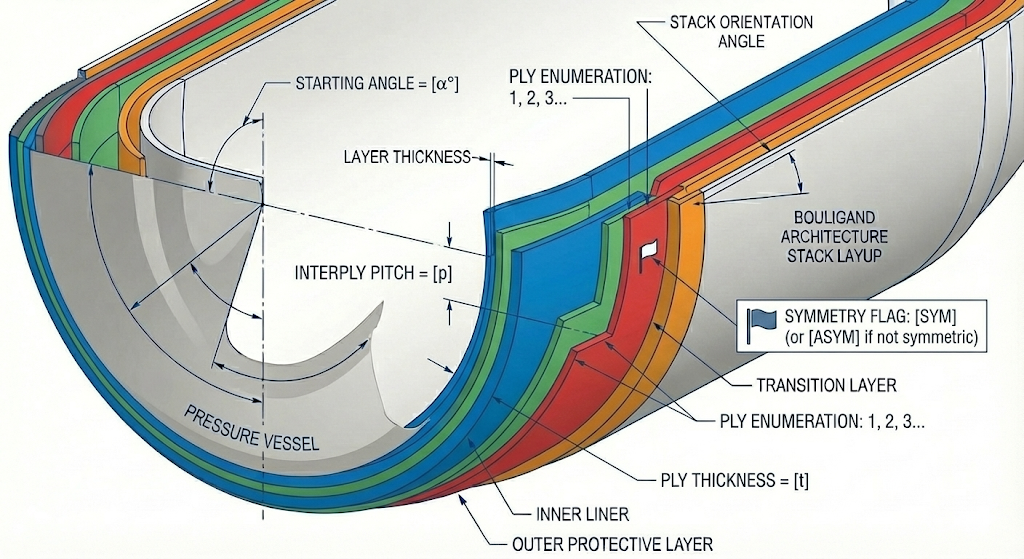}
    \caption{Schematic illustration of the bio-inspired Bouligand composite pressure-vessel architecture. The laminate is constructed from multiple plies
with gradually rotating fiber orientations, parameterized by a starting angle $\alpha$, inter-ply pitch $p$, ply enumeration through the thickness, ply
thickness $t$, and a symmetry flag controlling mid-plane mirroring. Together, these variables define the helicoidal stacking sequence used to generate the
pressure-vessel design space evaluated in this work.} 
    \label{fig:vessel}
\end{figure}

Figure~\ref{fig:vessel} schematically illustrates the Bouligand composite pressure-vessel architecture and highlights the key design parameters governing the helicoidal laminate stacking sequence, including ply orientation, inter-ply pitch, thickness, and symmetry.

\textbf{\textit{Simulation setup}}

For each configuration, a CFRP pressure vessel (cylindrical body with spherical end
caps; 165~mm diameter, 350~mm half-length) is simulated under an internal pressure of
62.5~MPa. Abaqus/CAE generates the geometry, symmetry boundary conditions, and S8R/STRI65
meshing. A unidirectional IM7/8551 lamina is modeled as an orthotropic ply with
Hashin damage initiation \cite{Hashin}. Field outputs include section-point stresses
$S_{11}$ and $S_{22}$, displacements, reaction forces, and failure flags.

\paragraph{Objective for BO.}
To compare surrogate models, we scalarize the multi-objective problem
(minimizing $S_{11}$, $S_{22}$, and thickness) into:
\begin{equation}
  \min J
  = 0.48\,\frac{S_{11}}{X_{t}}
  + 0.40\,\frac{S_{22}}{Y_{c}}
  + 0.12\,\text{Thick},
\end{equation}
where $(X_{t},Y_{c})=(\SI{2560}{\mega\pascal},\,\SI{185}{\mega\pascal})$ follow a
maximum-strength criterion \cite{Micromechanics}. Weighted-sum scalarization
\cite{Lee2019} is widely used to generate supported Pareto-optimal solutions;
here it provides a unified objective for BO-based surrogate comparison.

\subsubsection{Moir\'e Dataset}

\datasetstats{400{,}000}{6}{Jain's index}{maximize}

\begin{figure}[H]
    \centering
    \includegraphics[width=0.99\linewidth]{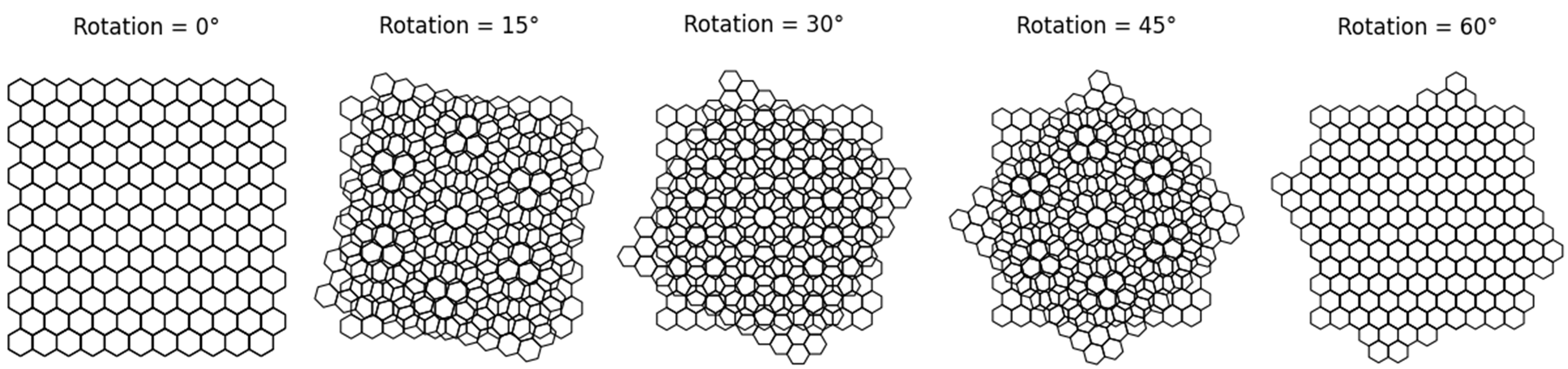}
    \caption{Moire.}
    \label{fig:Moire}
\end{figure}

\textbf{\textit{Dataset motivation and scope}}
Moir\'e patterns arise when two or more periodic meshes, grids, or lattices are superimposed with a slight mismatch in orientation, spacing, or alignment \cite{andrei2021marvels}. Figure~\ref{fig:Moire} shows the result of rotation of one layer over another, for six (6) rotation angles.  Even if each individual lattice is perfectly regular, their overlaid configuration generates a new, larger-scale interference pattern composed of slowly varying regions of local alignment and misalignment \cite{chen2020moire}. Small angular offsets or spacing differences accumulate over distance, giving rise to extended “supercells’’ whose characteristic length scale can be orders of magnitude larger than the primitive lattice \cite{shi2021exotic}.

In modern materials research, Moir\'e patterns serve both as a sensitive probe and a tunable design parameter \cite{PAPIA2026114270}. Their effects span natural and engineered systems. In van der Waals heterostructures—such as rotated bilayer graphene, transition-metal dichalcogenides, and related 2D materials—Moir\'e superlattices can radically reshape the electronic band structure, enabling phenomena central to twistronics and quantum devices \cite{jadaun2023review}. Beyond electronics, Moir\'e geometry can tailor transport and mechanical response in porous membranes \cite{pryds2024twisted}, photonic structures \cite{hu2021tailoring}, and magnetic materials \cite{song2021direct}.

\paragraph{Design space}
For each independent sample in our dataset, a multilayer hexagonal mesh was constructed using the \texttt{hexalattice} library (https://github.com/alexkaz2/hexalattice). In our study, we employed \textbf{six layers}. The first layer was fixed at a rotation of $0^\circ$, while the remaining five layers were assigned random rotation angles drawn uniformly from $0^\circ$ to $60^\circ$. All hexagons were rendered without fill, with a minimum hexagon diameter of one unit.

A centrally located window defined the fixed field of view. After rendering, each multilayer structure was rasterized to an RGBA image buffer, cropped, converted to grayscale, and thresholded using Otsu’s global method to isolate mesh linework. Connected components were then identified via standard region-labeling procedures. Only components fully contained within the cropped frame were retained, excluding any touching the image boundary. For each sample, the areas of all valid connected components (i.e., pores) were computed, and a histogram of pore areas was constructed using linearly spaced bins from zero to the maximum pore area observed in that sample.

Using this procedure, we generated a total of \textbf{400{,}000 samples}, each corresponding to a unique multilayer configuration and its resulting Moir\'e pore-area distribution. This large-scale dataset provides a high-dimensional, highly nonlinear, and geometrically complex landscape suitable for benchmarking surrogate-based Bayesian Optimization.

\paragraph{Objective for BO.}
For every pore-area distribution, we compute \emph{Jain's fairness index}, a classical measure of distributional uniformity:
\[
J = \frac{\left(\sum_i a_i\right)^2}{n\,\sum_i a_i^2},
\]
where $a_i$ are the pore areas and $n$ is the number of pores. Values close to~1 indicate highly uniform pore sizes, while lower values denote strong heterogeneity. This scalar serves as the BO target, quantifying how rotation-induced Moir\'e interference influences geometric regularity.

\subsubsection{Multi-PZT Semi-Active Tuned Mass Damper (SATMD)}

\datasetstats{200{,}000}{7}{multi-modal vibration}{minimize}

Semi-active tuned mass dampers (SATMDs) are widely used for vibration mitigation in large, lightweight, and flexible structures~\cite{Chatziathanasiou2022,Chatziathanasiou2022_2,Chatziathanasiou2025}. An SATMD combines an auxiliary mass with a piezoelectric element connected to a shunted RL network, enabling real-time modification of the system’s effective stiffness and damping properties. By appropriately tuning the resistance and inductance values of the shunt circuit, the device can suppress vibration amplitudes within targeted frequency bands. Determining the optimal tuning parameters, however, is non-trivial, as the SATMD dynamics depend simultaneously on the modal characteristics of the host structure and on the properties of the external excitation.

In multi-PZT SATMD configurations, multiple piezoelectric elements are incorporated into the same device, with each element equipped with its own independent shunt circuit. This arrangement introduces multiple anti-resonances into the host system, enhances multi-modal vibration attenuation, and enables effective vibration suppression over a broader frequency range. At the same time, the dimensionality of the tuning problem increases substantially, as multiple resistance and inductance parameters must be optimized concurrently, rendering exhaustive search strategies impractical.

To construct the dataset used in this work, a multi-PZT SATMD configuration was evaluated on a representative test structure exhibiting four dominant structural modes at approximately 36\,Hz, 47\,Hz, 65\,Hz, and 159\,Hz, as identified from the frequency response function (FRF) of the measured vertical acceleration at the fuselage tip. The SATMD comprised three piezoelectric elements, each connected to its own RL shunt circuit, while an auxiliary mass contributed a fourth anti-resonance. The resulting optimization variables are the auxiliary mass ratio $m_d$—defined as the ratio between the damper mass and a reference mass of the host structure—and the six shunt-circuit parameters $(R_1, R_2, R_3, L_1, L_2, L_3)$ associated with the three piezoelectric elements, yielding a seven-dimensional design space.

\begin{figure}[H]
    \centering
    \includegraphics[width=0.99\linewidth]{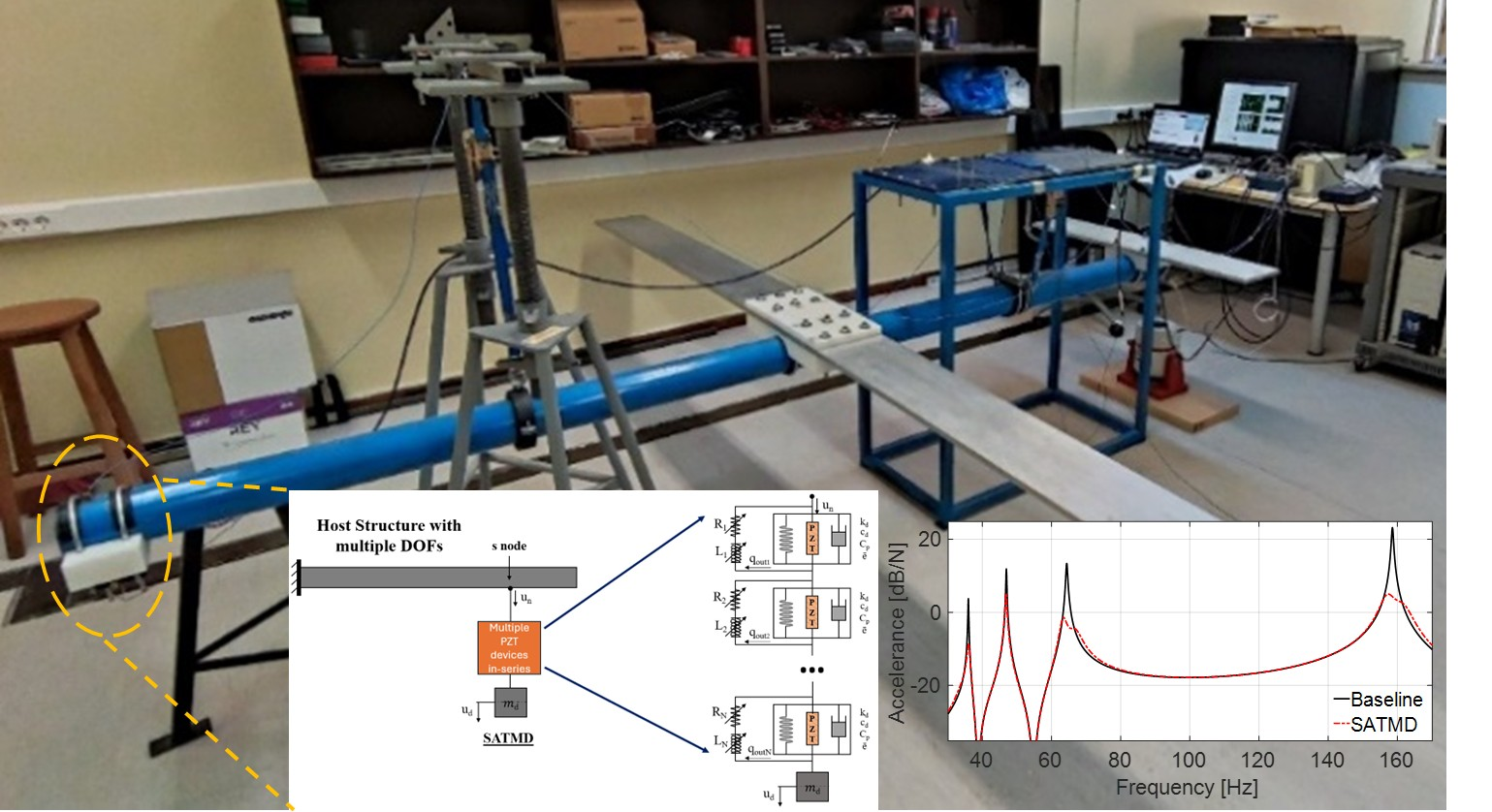}
    \caption{Photograph of the experimental test rig used to evaluate the multi-PZT SATMD mounted on a flexible host structure (center), with the damper location highlighted. A schematic representation of the multi-PZT SATMD configuration is shown (inset), illustrating the auxiliary mass and multiple piezoelectric elements connected in series, each equipped with an independent shunt circuit. Frequency response functions of the vertical acceleration at the measurement location are shown for the baseline structure (solid black) and for a representative SATMD configuration (dashed red), demonstrating simultaneous attenuation of the four dominant structural modes within the frequency bands targeted by the BO objective.}
    \label{fig:PZT}
\end{figure}

\paragraph{Objective for Bayesian Optimization.}
The objective of the optimization is to suppress the system response within frequency regions where the vibration magnitude lies at least 20\,dB below the modal peaks. Figure~\ref{fig:PZT} illustrates this objective by comparing the baseline frequency response of the host structure, characterized by four dominant structural modes, with the attenuated response obtained using a multi-PZT SATMD for a representative configuration of the auxiliary mass ratio and shunt-circuit parameters.

Since lower-frequency modes typically induce larger vibration amplitudes, frequency-dependent weighting factors were introduced to emphasize mitigation in the most critical bands. The optimal SATMD parameters are obtained by minimizing the following weighted objective function:
\begin{equation}
\begin{aligned}
\min \; f\!\left(\mathrm{SATMD}_{m_d,R_1,L_1,R_2,L_2,R_3,L_3}\right)
&=
w_1 \int_{35\,\mathrm{Hz}}^{37\,\mathrm{Hz}}
\left|\ddot{u}_{\mathrm{tip,SATMD}}(f)\right|\,\mathrm{d}f \\
&\quad +
w_2 \int_{46\,\mathrm{Hz}}^{48\,\mathrm{Hz}}
\left|\ddot{u}_{\mathrm{tip,SATMD}}(f)\right|\,\mathrm{d}f \\
&\quad +
w_3 \int_{62\,\mathrm{Hz}}^{68\,\mathrm{Hz}}
\left|\ddot{u}_{\mathrm{tip,SATMD}}(f)\right|\,\mathrm{d}f \\
&\quad +
w_4 \int_{156\,\mathrm{Hz}}^{162\,\mathrm{Hz}}
\left|\ddot{u}_{\mathrm{tip,SATMD}}(f)\right|\,\mathrm{d}f .
\end{aligned}
\end{equation}

A total of $2\times10^{5}$ distinct SATMD configurations were generated within the prescribed parameter bounds. For each configuration, vibration attenuation was evaluated in the vicinity of the four structural modes, and the weighted objective value was computed accordingly.

% \section*{Declarations}
\backmatter
\bmhead{Acknowledgements}
This work was supported by the European Union’s Horizon Europe research and innovation programme under grant agreement No. 101135927 (NOUS).

\bmhead{Supporting Information}
Supporting Information accompanies this publication.

\bmhead{Data availability}
Datasets are deposited at \url{10.5281/zenodo.18194022}.

\bmhead{Code availability}
The BO and evaluation metrics estimation codes underlying this work are freely available for general use under Apache 2.9 LICENSE (\url{https://www.apache.org/licenses/LICENSE-2.0}) and are deposited at \url{https://github.com/insane-group/FruBO/}.

\bmhead{Declarations}
The authors declare no competing interests.

\bigskip
\begin{flushleft}%
Editorial Policies for:

\bigskip\noindent
Springer journals and proceedings: \url{https://www.springer.com/gp/editorial-policies}

\bigskip\noindent
Nature Portfolio journals: \url{https://www.nature.com/nature-research/editorial-policies}

\bigskip\noindent
\textit{Scientific Reports}: \url{https://www.nature.com/srep/journal-policies/editorial-policies}

\bigskip\noindent
BMC journals: \url{https://www.biomedcentral.com/getpublished/editorial-policies}
\end{flushleft}

\begin{appendices}

\section{Section title of first appendix}\label{secA1}

An appendix contains supplementary information that is not an essential part of the text itself but which may be helpful in providing a more comprehensive understanding of the research problem or it is information that is too cumbersome to be included in the body of the paper.




\end{appendices}


\bibliography{sn-bibliography}% common bib file

\end{document}

\title[Article Title]{Frugal Bayesian Optimization: Scalable Surrogates for Data- and Resource-Limited Discovery}


\author*[1]{\fnm{Panagiotis} \sur{Krokidas}}\email{p.krokidas@iit.demokritos.gr}

\author[1,2]{\fnm{Christoforos} \sur{Rekatsinas}}\email{crek@iit.demokritos.gr}
\author[1,3]{\fnm{Vassilis} \sur{Sioros}}\email{vsioros@iit.demokritos.gr}

\author[4]{\fnm{Grigorios M.} \sur{Chatziathanasiou}}\email{grigorischatz@hmu.gr}

\author[5,6]{\fnm{Efi-Maria} \sur{Papia}}\email{e.papia@inn.demokritos.gr}

\author[1,7]{\fnm{George} \sur{Giannakopoulos}}\email{ggianna@iit.demokritos.gr}

\affil*[1]{\orgdiv{Institute of Informatics and Telecommunications}, \orgname{National Centre for Scientific Research "Demokritos"}, \orgaddress{
% \street{},
\city{Agia Paraskevi},
% \postcode{}, \state{},
\country{Greece}}}

\affil[2]{\orgdiv{Department of Mechanical Engineering and Aeronautics}, \orgname{University of Patras}, \orgaddress{\city{Patras} \country{Greece}}}

\affil[3]{\orgdiv{Department of Informatics and Telecommunications}, \orgname{National and Kapodistrian University of Athens}, \orgaddress{ \city{Athens}, \country{Greece}}}

\affil[4]{\orgdiv{School of Mechanical Engineering}, \orgname{Hellenic Mediterranean
University}, \orgaddress{\city{Heraklion}, \state{Crete}, \country{Greece}}}

\affil*[5]{\orgdiv{Institute of Nanoscience and Nanotechnology}, \orgname{National Centre for Scientific Research "Demokritos"}, \orgaddress{\city{Agia Paraskevi}, \country{Greece}}}

\affil[6]{\orgdiv{Department of Physics}, \orgname{National and Kapodistrian University of Athens}, \orgaddress{\city{Athens}, \country{Greece}}}

\affil[7]{\orgname{SciFY PNPC}, \orgaddress{\city{Agia Paraskevi}, \country{Greece}}}

% \keywords{keyword1, Keyword2, Keyword3, Keyword4}



\maketitle

\section{The Benchmarking Functions of this work}

\begin{figure}[H]
    \centering
    \includegraphics[width=0.65\linewidth]{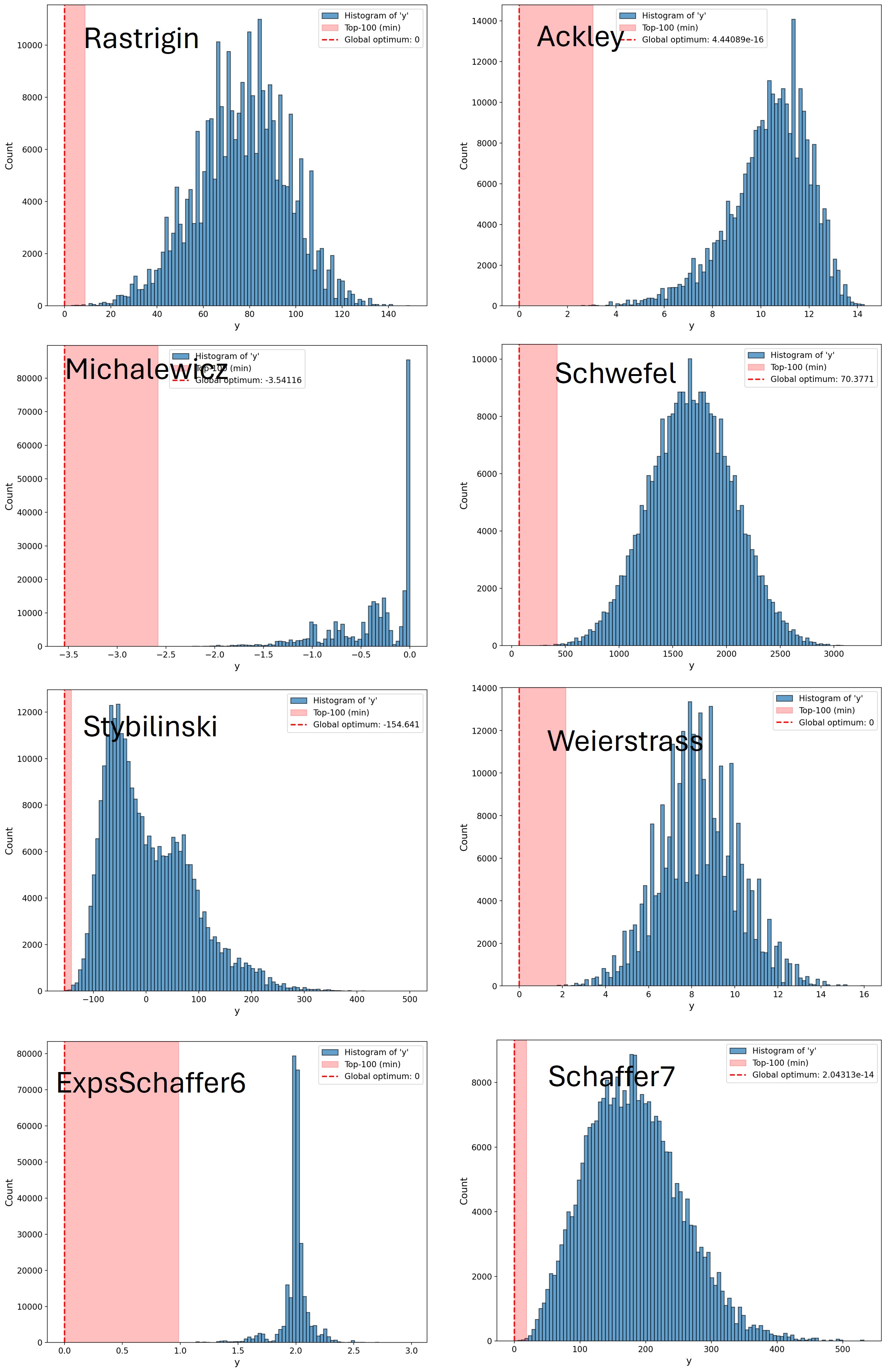}
    \caption{distribution plots for the target property, showing the global optimum and the top-100 values}
    \label{fig:funct_distributions}
\end{figure}

\begin{figure}[H]
    \centering
    \includegraphics[width=0.95\linewidth]{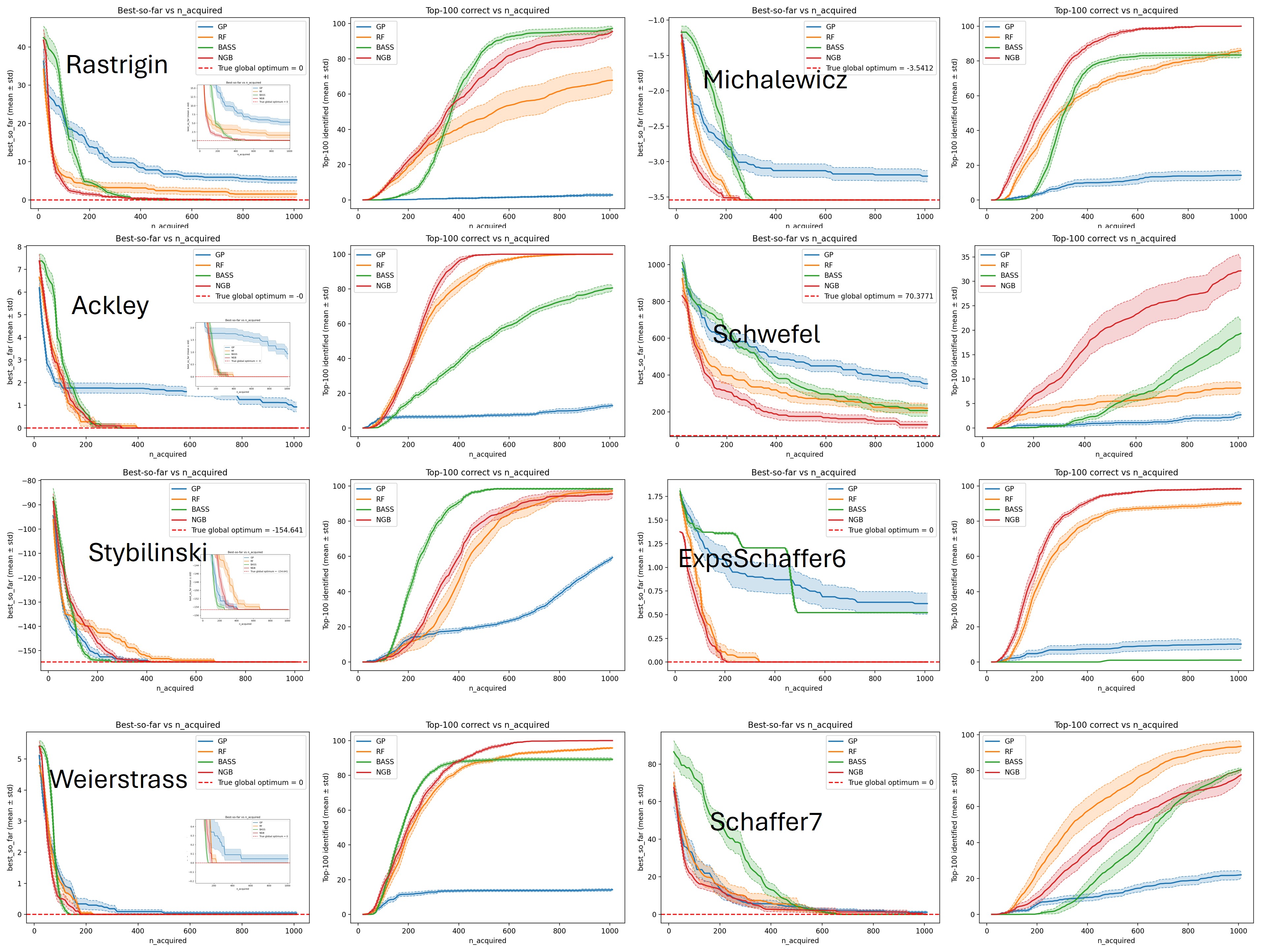}
    \caption{optimization results in terms of best solution and top-100 solutions identification as a function of sample acquisition}
    \label{fig:funct_AllPerformancePlots}
\end{figure}

\section{AUC areas}
% -------- BASS --------
\begin{table}[htbp!]
    \centering
    \caption{AUC metrics for the BASS surrogate across benchmark functions.}
    \label{tab:auc_bass}
    \small
    \begin{tabular}{lcccccc}
        \toprule
        & \multicolumn{3}{c}{AUC over time} & \multicolumn{3}{c}{AUC over samples} \\
        \cmidrule(lr){2-4} \cmidrule(lr){5-7}
        Dataset & Best found & Top--100 acquired & Product & Best found & Top--100 acquired & Product \\
        \midrule
        expschaffer6 & 6.482  & 742.2   & 4.81E+03 & 871.3    & 9.85E+04 & 8.58E+07 \\
        rastrigin    & 31.4   & 339.9   & 1.07E+04 & 4411     & 3.82E+04 & 1.68E+08 \\
        Michalewicz  & 2.486  & 313.1   & 7.78E+02 & 366.1    & 3.99E+04 & 1.46E+07 \\
        Ackley       & 4.927  & 578.5   & 2.85E+03 & 693      & 5.34E+04 & 3.70E+07 \\
        Schwefel     & 2907   & 6.83E+02 & 1.99E+06 & 4.10E+05 & 9.28E+04 & 3.81E+10 \\
        Styblinski   & 58.15  & 210.8   & 1.23E+04 & 5953     & 2.33E+04 & 1.38E+08 \\
        Weierstrass  & 2.013  & 226.8   & 4.57E+02 & 301.2    & 2.47E+04 & 7.45E+06 \\
        Schaffer7    & 135    & 5.20E+02 & 7.02E+04 & 1.96E+04 & 6.63E+04 & 1.30E+09 \\
        \bottomrule
    \end{tabular}
\end{table}

% -------- GP --------
\begin{table}[htbp!]
    \centering
    \caption{AUC metrics for the GP surrogate across benchmark functions.}
    \label{tab:auc_gp}
    \small
    \begin{tabular}{lcccccc}
        \toprule
        & \multicolumn{3}{c}{AUC over time} & \multicolumn{3}{c}{AUC over samples} \\
        \cmidrule(lr){2-4} \cmidrule(lr){5-7}
        Dataset & Best found & Top--100 acquired & Product & Best found & Top--100 acquired & Product \\
        \midrule
        expschaffer6 & 6.278  & 827.4   & 5.19E+03 & 846.8   & 9.17E+04 & 7.76E+07 \\
        rastrigin    & 55.45  & 856.6   & 4.75E+04 & 9748    & 9.77E+04 & 9.52E+08 \\
        Michalewicz  & 3.591  & 788.7   & 2.83E+03 & 559     & 3.99E+04 & 2.23E+07 \\
        Ackley       & 12.18  & 806.1   & 9.82E+03 & 1650    & 9.14E+04 & 1.51E+08 \\
        Schwefel     & 4333   & 1.03E+03 & 4.47E+06 & 4.99E+05 & 9.79E+04 & 4.89E+10 \\
        Styblinski   & 24     & 592.4   & 1.42E+04 & 5423    & 7.46E+04 & 4.04E+08 \\
        Weierstrass  & 0.7901 & 775.4   & 6.13E+02 & 348.6   & 8.69E+04 & 3.03E+07 \\
        Schaffer7    & 24.86  & 7.37E+02 & 1.83E+04 & 8523    & 8.70E+04 & 7.42E+08 \\
        \bottomrule
    \end{tabular}
\end{table}

% -------- NGBoost --------
\begin{table}[htbp!]
    \centering
    \caption{AUC metrics for the NGBoost surrogate across benchmark functions.}
    \label{tab:auc_ngb}
    \small
    \begin{tabular}{lcccccc}
        \toprule
        & \multicolumn{3}{c}{AUC over time} & \multicolumn{3}{c}{AUC over samples} \\
        \cmidrule(lr){2-4} \cmidrule(lr){5-7}
        Dataset & Best found & Top--100 acquired & Product & Best found & Top--100 acquired & Product \\
        \midrule
        expschaffer6 & 0.8508 & 188     & 1.60E+02 & 96.79   & 2.04E+04 & 1.97E+06 \\
        rastrigin    & 15.41  & 359.1   & 5.53E+03 & 1791    & 3.90E+04 & 6.98E+07 \\
        Michalewicz  & 0.7943 & 188.8   & 1.50E+02 & 96.63   & 2.14E+04 & 2.07E+06 \\
        Ackley       & 4.273  & 193.7   & 8.28E+02 & 494.9   & 2.15E+04 & 1.06E+07 \\
        Schwefel     & 1803   & 6.35E+02 & 1.14E+06 & 2.33E+05 & 8.11E+04 & 1.89E+10 \\
        Styblinski   & 63.9   & 341     & 2.18E+04 & 7223    & 3.77E+04 & 2.73E+08 \\
        Weierstrass  & 1.7    & 182.3   & 3.10E+02 & 212.7   & 8.69E+04 & 1.85E+07 \\
        Schaffer7    & 60.42  & 529.8   & 3.20E+04 & 6788    & 5.74E+04 & 3.89E+08 \\
        \bottomrule
    \end{tabular}
\end{table}

% -------- Random Forest --------
\begin{table}[htbp!]
    \centering
    \caption{AUC metrics for the Random Forest surrogate across benchmark functions.}
    \label{tab:auc_rf}
    \small
    \begin{tabular}{lcccccc}
        \toprule
        & \multicolumn{3}{c}{AUC over time} & \multicolumn{3}{c}{AUC over samples} \\
        \cmidrule(lr){2-4} \cmidrule(lr){5-7}
        Dataset & Best found & Top--100 acquired & Product & Best found & Top--100 acquired & Product \\
        \midrule
        expschaffer6 & 0.6056 & 152.1   & 9.21E+01 & 139.3   & 3.04E+04 & 4.23E+06 \\
        rastrigin    & 16.68  & 275.8   & 4.60E+03 & 3536    & 5.68E+04 & 2.01E+08 \\
        Michalewicz  & 0.6494 & 194.7   & 1.26E+02 & 158     & 4.03E+04 & 6.37E+06 \\
        Ackley       & 2.008  & 117.9   & 2.37E+02 & 470.9   & 2.48E+04 & 1.17E+07 \\
        Schwefel     & 1808   & 5.59E+02 & 1.01E+06 & 3.23E+05 & 9.39E+04 & 3.03E+10 \\
        Styblinski   & 35.63  & 195.2   & 6.95E+03 & 7737    & 4.18E+04 & 3.23E+08 \\
        Weierstrass  & 1.214  & 136.3   & 1.65E+02 & 264.1   & 2.62E+04 & 6.92E+06 \\
        Schaffer7    & 36.66  & 184.2   & 6.75E+03 & 8403    & 4.00E+04 & 3.36E+08 \\
        \bottomrule
    \end{tabular}
\end{table}

\begin{figure}[H]
    \centering
    \includegraphics[width=0.8\linewidth]{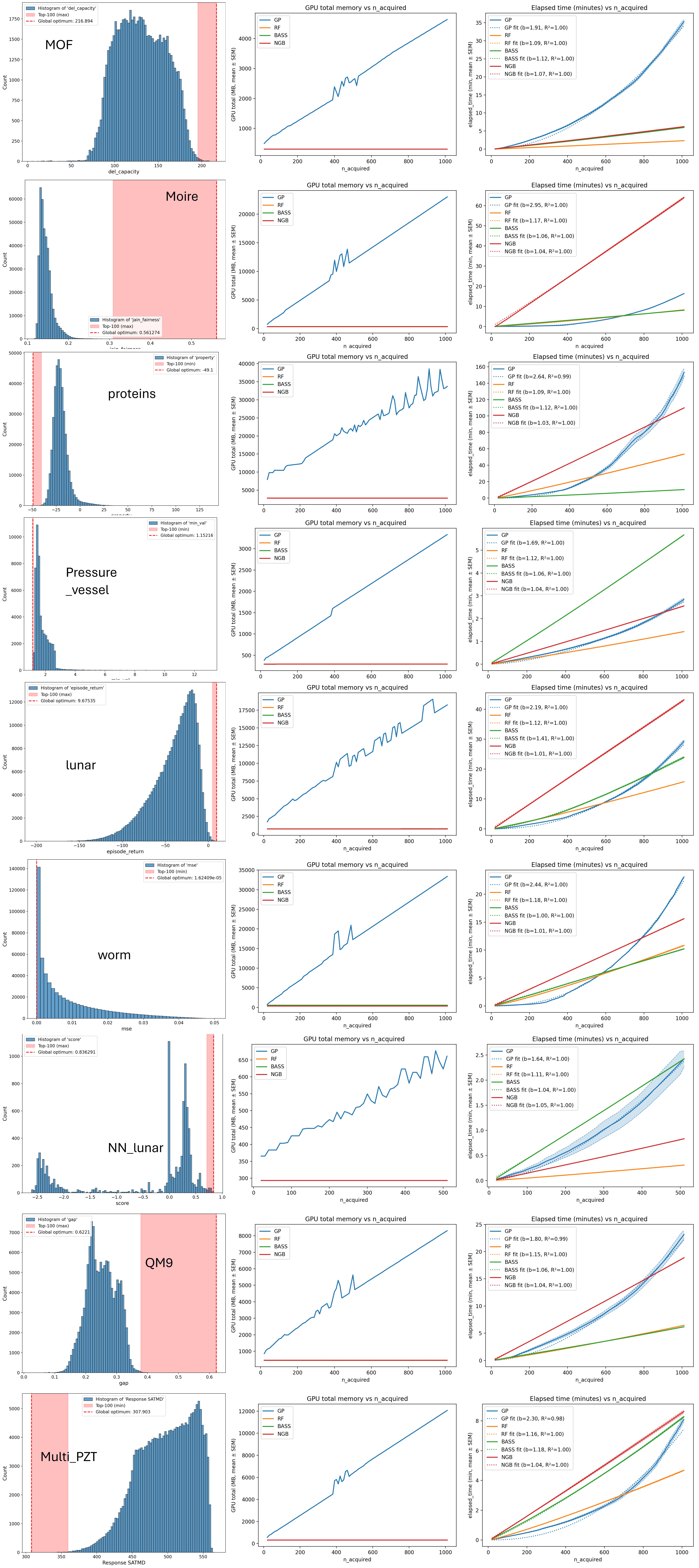}
    \caption{distribution plots for the target property, showing the global optimum and the top-100 values}
    \label{fig:realCases_Distr_gpus_clocltime}
\end{figure}

% -------- BASS --------
\begin{table}[htbp!]
    \centering
    \caption{AUC metrics for the BASS surrogate across real-case datasets.}
    \label{tab:auc_bass_real}
    \small
    \begin{tabular}{lcccccc}
        \toprule
        & \multicolumn{3}{c}{AUC over time} & \multicolumn{3}{c}{AUC over samples} \\
        \cmidrule(lr){2-4} \cmidrule(lr){5-7}
        Dataset & Best found & Top--100 acquired & Product & Best found & Top--100 acquired & Product \\
        \midrule
        Moire           & 0.2773    & 678.5  & 1.88E+02 & 34.38  & 8.40E+04 & 2.89E+06 \\
        pressure\_vessel& 0.1666    & 536.2  & 8.93E+01 & 29.93  & 9.47E+04 & 2.84E+06 \\
        MOFs            & 13.15     & 289.7  & 3.81E+03 & 2424   & 4.98E+04 & 1.21E+08 \\
        proteins        & 61.2      & 979.2  & 5.99E+04 & 6047   & 9.47E+04 & 5.73E+08 \\
        lunar           & 187.4     & 2372   & 4.45E+05 & 8768   & 9.88E+04 & 8.66E+08 \\
        worm            & 5.353E-04 & 1010   & 5.41E-01 & 0.05232& 9.90E+04 & 5.18E+03 \\
        neural          & 0.08093   & 204.6  & 1.66E+01 & 16.83  & 4.23E+04 & 7.12E+05 \\
        QM9            & 0.05715   & 297.6  & 1.54E+01 &  8.26  & 4.87E+04 & 4.02E+05 \\
        multi\_PZT     & 44.81   & 662.9  & 2.97E+04 &  5591  & 8.16E+04 & 4.56E+08 \\
        \bottomrule
    \end{tabular}
\end{table}

% -------- GP --------
\begin{table}[htbp!]
    \centering
    \caption{AUC metrics for the GP surrogate across real-case datasets.}
    \label{tab:auc_gp_real}
    \small
    \begin{tabular}{lcccccc}
        \toprule
        & \multicolumn{3}{c}{AUC over time} & \multicolumn{3}{c}{AUC over samples} \\
        \cmidrule(lr){2-4} \cmidrule(lr){5-7}
        Dataset & Best found & Top--100 acquired & Product & Best found & Top--100 acquired & Product \\
        \midrule
        Moire           & 0.08042   & 1291   & 1.04E+02 & 38.27   & 8.42E+04 & 3.22E+06 \\
        pressure\_vessel& 0.05793   & 259    & 1.50E+01 & 25.46   & 9.24E+04 & 2.35E+06 \\
        MOFs            & 43.95     & 1612   & 7.08E+04 & 3024    & 5.41E+04 & 1.64E+08 \\
        proteins        & 754.5     & 1.39E+04 & 1.05E+07 & 5996   & 9.27E+04 & 5.56E+08 \\
        lunar           & 117.9     & 2887   & 3.40E+05 & 5360    & 9.81E+04 & 5.26E+08 \\
        worm            & 6.54E-06  & 2078   & 1.36E-02 & 5.353E-04 & 9.26E+04 & 4.96E+01 \\
        neural          & 0.03758   & 72.85  & 2.74E+00 & 12.81   & 2.01E+04 & 2.57E+05 \\
        GM9             & 0.209     & 1719  & 3.59E+02 & 24.73   & 7.63E+04 & 1.89E+06 \\
        multi\_PZT       & 43.14     & 423.6  & 1.83E+04 & 5828   & 6.32E+04 & 3.69E+08 \\
        \bottomrule
    \end{tabular}
\end{table}

% -------- NGBoost --------
\begin{table}[htbp!]
    \centering
    \caption{AUC metrics for the NGBoost surrogate across real-case datasets.}
    \label{tab:auc_ngb_real}
    \small
    \begin{tabular}{lcccccc}
        \toprule
        & \multicolumn{3}{c}{AUC over time} & \multicolumn{3}{c}{AUC over samples} \\
        \cmidrule(lr){2-4} \cmidrule(lr){5-7}
        Dataset & Best found & Top--100 acquired & Product & Best found & Top--100 acquired & Product \\
        \midrule
        Moire            & 0.6535   & 1082    & 7.07E+02 & 52.34  & 8.79E+04 & 4.60E+06 \\
        pressure\_vessel & 0.04698  & 219.5   & 1.03E+01 & 18.52  & 8.60E+04 & 1.59E+06 \\
        MOFs             & 32.03    & 221.8   & 7.10E+03 & 5268   & 3.66E+04 & 1.93E+08 \\
        proteins         & 632.5    & 1.03E+04& 6.53E+06 & 5787   & 9.39E+04 & 5.43E+08 \\
        lunar            & 181.1    & 4208    & 7.62E+05 & 4197   & 9.75E+04 & 4.09E+08 \\
        worm             & 7.51E-05 & 1531    & 1.15E-01 & 7814   & 9.79E+04 & 7.65E+08 \\
        neural           & 0.02141  & 29.92   & 6.41E-01 & 12.85  & 1.81E+04 & 2.32E+05 \\
        QM9              & 1.505    & 799.2   & 1.20E+03 & 81.13  & 4.28E+04 & 3.47E+06 \\
        multi\_PZT       & 49.65    & 437.4   & 2.17E+04 & 5764   & 5.10E+04 & 2.94E+08 \\
        \bottomrule
    \end{tabular}
\end{table}

% -------- Random Forest --------
\begin{table}[htbp!]
    \centering
    \caption{AUC metrics for the Random Forest surrogate across real-case datasets.}
    \label{tab:auc_rf_real}
    \small
    \begin{tabular}{lcccccc}
        \toprule
        & \multicolumn{3}{c}{AUC over time} & \multicolumn{3}{c}{AUC over samples} \\
        \cmidrule(lr){2-4} \cmidrule(lr){5-7}
        Dataset & Best found & Top--100 acquired & Product & Best found & Top--100 acquired & Product \\
        \midrule
        Moire           & 0.307    & 742.2  & 2.28E+02 & 41.63  & 8.76E+04 & 3.65E+06 \\
        pressure\_vessel& 0.03322  & 129.9  & 4.32E+00 & 24.73  & 9.08E+04 & 2.25E+06 \\
        MOFs            & 4.482    & 99.23  & 4.45E+02 & 2111   & 4.38E+04 & 9.24E+07 \\
        proteins        & 303      & 5068   & 1.54E+06 & 5768   & 9.46E+04 & 5.45E+08 \\
        lunar           & 99.66    & 1556   & 1.55E+05 & 7495   & 9.83E+04 & 7.37E+08 \\
        worm            & 3.31E-05 & 1072   & 3.55E-02 & 5029   & 9.85E+04 & 4.96E+08 \\
        neural          & 0.02688  & 23.37  & 6.28E-01 & 43.77  & 3.78E+04 & 1.66E+06 \\
        QM9             & 0.08900  & 435.40  & 3.88E+01 & 17.20 & 6.81E+04 & 1.17E+06 \\
        multi\_PZT       & 19.22  & 418.40  & 8.04E+03 & 4336 & 8.96E+04 & 3.89E+08 \\
        \bottomrule
    \end{tabular}
\end{table}
\FloatBarrier

\section{Computational Setup}
All computations were performed on a personal desktop equipped with an Intel Core i9-10900K CPU, 64 GB of RAM, and an NVIDIA GeForce RTX 3070 Ti GPU (8 GB VRAM), running Windows 11 Pro (64-bit).


% No bibliography is needed in the SI because it contains no citation commands.